\documentclass[final,12pt]{clear2025full} 

\usepackage{ZijunMacro}
\usepackage{graph-separation}
\usepackage{tikz-cd}
\usepackage{array}

\title[Counterfactual explainability]{Counterfactual Explainability
  and Analysis of Variance}
\usepackage{times}

\clearauthor{%
  \Name{Zijun Gao} \Email{zijungao@marshall.usc.edu}\\
  \addr Department of Data Science and Operations, University of Southern California
  \AND
  \Name{Qingyuan Zhao} \Email{qyzhao@statslab.cam.ac.uk}\\
  \addr Statistical Laboratory, University of Cambridge
}

\begin{document}

\maketitle

\begin{abstract}
  Existing tools for explaining complex models and systems are
  associational rather than causal and do not provide mechanistic
  understanding. We propose a new notion called counterfactual
  explainability for causal attribution that is motivated by the
  concept of genetic heritability in twin studies. Counterfactual
  explainability extends methods for global sensitivity analysis
  (including the functional analysis of variance and Sobol's indices),
  which assumes independent explanatory variables, to dependent
  explanations by using a directed acyclic graphs to describe their
  causal relationship. Therefore, this explanability measure
  directly incorporates causal mechanisms by construction. Under a
  comonotonicity assumption, we discuss methods for estimating
  counterfactual explainability and apply them to a real dataset
  dataset to explain income inequality by gender, race, and
  educational attainment.
\end{abstract}

\begin{keywords}
  Causality, Directed acyclic graph, Explainable AI, Genetic
  heritability, Global sensitivity analysis.
\end{keywords}

\section{Introduction}

Explanation and causal attribution are at the core of human reasoning
and scientific inquisition. One good example for this is the concept
of heritability in the long-standing debate about nature versus
nurture in biological and social sciences, dating back at least to
\citet{galtonRegressionMediocrityHereditary1886}. Heuristically,
heritability should meansure the proportion of phenotypic
variation due to genetic variation, but there is much confusion and
debate about various definitions of heritability
\citep{jacquardHeritabilityOneWord1983,zaitlenHeritabilityGenomewideAssociation2012,yangConceptsEstimationInterpretation2017}. Part
of the difficulty in interpreting heritability estimates
(especially those based on unrelated inviduals) is that they usually
only model statistical associations, but heritability is an
inherent causal concept. Causal explanations are also routinely sought
in many other disciplines, including
earth sciences \citep{rungeInferringCausationTime2019}, social science
\citep{vanderweele2015explanation}, reliability engineering
\citep{ioossReviewGlobalSensitivity2015}, and law
\citep{dawidEffectsCausesCauses2022}.

With wider applications of increasingly more complex machine learning
algorithms, there is also a rapidly growing need to explain black-box
prediction models.
There is a
long-standing and quickly expanding literature on interpretable
summaries and visualization tools for black-box models, including
Sobol's indices in global
sensitivity analysis \citep{sobol1993sensitivity}, functional analysis
of variance \citep{hoeffding1948}, Shapley's value in game theory
\citep{shapley1953value}, partial dependence plots
\citep{friedman2001greedy}, accumulated local effect plots
\citep{apley2020visualizing}, and various other variable importance
 measures \citep{breiman2001random, hines2022variable,
  williamson2023general} and generalized correlation measures
\citep{zhengGeneralizedMeasuresCorrelation2012,allenGeneralizedMeasuresCorrelation2022}.
However, most existing methods for ``explainable
artificial intelligence'', though with links to causality
\citep{zhao2021causal}, are ultimately associational. Thus, a main
drawback of these tools is that they cannot incorporate mechanistic
knowledge or other contextual information that is crucial
for model explanation in real world problems.
Relatedly, a widely held view in the machine learning
community is that some considerations of the causal (and perhaps
counterfactual) relationship between model inputs are
necessary in formulating good notions of fairness
\citep{kusner2017counterfactual,imai2023principal}. In
a recent survey article for fairness in machine learning, it is
concluded that ``tools potentially
based on causal methods are needed to assist in the identification of protected
variables and groups as well as their proxies'' to make machine
learning ``fairer'' \citep{caton2024fairness}.

\subsection{Why explanations need to be causal?}
\label{sec:why-expl-need}

We give two simple examples to illustrate why a lack of causal
interpretation can be problematic. For the rest
of this paper, we consider how a real-valued outcome
variable $Y$ can be explained by some explanatory variables
$W_1,\dots,W_K$ ($K \geq 1$). In black-box prediction models, $Y$ is
often a deterministic function of $W = (W_1,\dots,W_K)$, but $Y$ can be an
arbitrary random variable in general (such as a human trait in the
heritability problem). We will use potential outcomes to
distinguish between statistical and causal models
\citep{imbens2015causal}; for example, we will use $Y(w_1,\dots,w_K)$
to denote the potential outcome of $Y$ in an intervention that sets
$W_1 = w_1, \dots, W_K = w_K$, and the realized outcome is given by $Y
= Y(W_1, \dots, W_K)$.

\begin{example}
  \label{exam:causally.independent}
  Let $ W_1 $ and $ W_2 $ be Rademacher random variables (so they
  take $-1$ or $1$ with probability a half), and suppose they are
  \emph{causally independent}, meaning they are not only independent
  in a probabilistic sense but also have no causal effect on each
  other. Suppose the outcome is generated by the causal model
  \[
    Y(w_1,w_2) = w_1 w_2.
  \]
  Thus, $W_1$ does have some causal influence on $Y$: we have, by
  recursive substitution of potential outcomes (see
  Definition \ref{def:causal-markov} below),
  \[
    Y(w_1) - Y(w_1') = Y(w_1, W_2) - Y(w_1', W_2) = (w_1 - w_1') W_2
    \neq 0 \quad \text{in general}.
  \]
  However, if $W_2$ is not observed, using just the probability
  distribution of $(W_1,Y)$ one will likely conclude that $W_1$
  offers no explanation to $Y$ at all because $W_1$ and $Y$ are
  statistically independent (even though they are not causally
  independent).
\end{example}

\begin{example}
  \label{exam:copy.cat}
  Suppose $W_1 = W_2$ with probability $1$, so it may seem sensible to
  divide explainability to some outcome $Y$ equally between $W_1$ and
  $W_2$. However, we often know why $W_1$ and $W_2$ are identical in
  the real world, and our explanations should take that into
  account. For example, if $W_1$ and $W_2$ represent the content of
  two identical videos in an online platform but the second video is
  simply an unauthorized copy of the first video (so $W_2(w_1) =
  w_1$), the credit should be attributed entirely to the original
  creator of $W_1$ and not at all to the copycat. Similarly, if $W_1$
  is the output of a weather forecast model and $W_2$ is the forecast
  reported by a weather presenter on television, it is the model, not
  the weather presenter, that is able to forecast weather accurately.
\end{example}

These examples demonstrate the need to consider the causal
relationship between the explanatory variables in order to provide
convincing explanations.

\subsection{Towards counterfactual explainability}
\label{sec:count-expl}

Counterfactual
causality (contrasting potential outcomes in ``different worlds'') is
sometimes regarded as the foundation for explanation
\citep{pearl2009,vanderweele2015explanation}; in view of Pearl's
Ladder of Causation (association-intervention-counterfactuals), it is
crucial to ``climb up'' to the counterfactual rung for ``imagining,
retrospection, understanding'' \citep{bookofwhy}. Counterfactual
reasoning can be found in causal mediation analysis and applications
to law and epidemiology; some key concepts include natural
direct/indirect effect
\citep{robinsIdentifiabilityExchangeabilityDirect1992,pearl2001uai},
probability of causation
\citep{robins1989probability,tian2000probabilities,dawid2017probability}, and notions
related to attributable fraction \citep{SuzukiYamamoto+2023}. However,
with a few exceptions
\citep{luEvaluatingCausesEffects2023,janzingQuantifyingIntrinsicCausal2024},
there is relatively little work on counterfactual explainability in
complex systems with many explanatory variables.

Our notion of counterfactual heritability is motivated by the
definition of heritability in twin studies \citep[see e.g.][p.\
434-441]{hartl1997principles}. Although the genetics literature
usually does not use a causal language explicitly (but see
\citet{kohler2011social} for an exception), heritability can be
intuitively defined by comparing the same trait (e.g., height) of
non-identical twins.
Let gene $G$ and environment $E$ be two causally independent factors
for some trait $Y = Y(G, E)$ and let $G$ denote the genotype of person
A. By ``causally independent'', we mean $G$ and $E$ are independent
random variables that have no causal effect on each other. Now
consider a hypothetical ``non-identical twin'' B of A that
has an independent and identically distributed (i.i.d.) draw of genes
denoted by $G'$ and goes through
exactly the same environment $E$. Counterfactual heritability can be
measured by how much A differs from B on average; formally,
\begin{equation}
  \label{eq:xi-heritability}
  \xi = \frac{\var(Y(G, E) - Y(G', E))}{2 \var(Y(G, E))} = 1 -
  \cor(Y(G, E), Y(G', E)).
\end{equation}
This can be viewed as a counterfactual, nonparametric
generalization of Falconer's formula in genetics which defines
heritability by comparing the correlations between identical and non-identical
twins \citep{falconerIntroductionQuantitativeGenetics1996}. See a
parallel work \citep{leiHeritabilityCounterfactualPerspective2025} for further
comparison with existing notions of heritability in genetics. Readers
who are familiar with global sensitivity analysis will recognize
\eqref{eq:xi-heritability} as the pick-freeze method for Sobol's upper
sensitivity index \citep{jansenAnalysisVarianceDesigns1999}, a
connection we will further explore below.

\subsection{Overview of this article}
\label{sec:overview}

The definition in \eqref{eq:xi-heritability} directly motivates a
notion of ``total
explainability'' when $Y$ depnds on several causally independent
factors $W_1,\dots,W_K$. In fact, when $Y$ is a deterministic function of
$W_1,\dots,W_K$, this definition coincides with Sobol's upper index in
global sensitivity analysis and \eqref{eq:xi-heritability} is
often called the ``pick-freeze'' method
\citep{sobol1993sensitivity}. To this end, we review the functional
analysis of variance and global sensitivity
analysis in \Cref{sec:literature-review}, show that several notions of
variable importance in the literature can be unified using what we
call the explanation algebra in \Cref{sec:unif-vari-import}, and
provide an axiomatic justification of
total explainability/Sobol's upper index in
\Cref{sec:axiom-total-expl}.

In \Cref{sec:independent}, we then propose a counterfactual
generalization to the global sensitivity analysis. Our approach is
based on defining total explainability using \eqref{eq:xi-heritability} but
with the variables replaced by their ``basic potential
outcomes'' (which can be thought as their intrinsic noise). This
allows us to define counterfactual explainability for
dependent explanatory variables $W_1,\dots,W_K$ if the causal
relationship between $W_1,\dots,W_K,Y$ can be described by a directed
acyclic graph (DAG). Compared to global sensitivity analysis, this more general
notion of explainability does not require the outcome variable $Y$ to
be a deterministic function of the explanatory variables and have
a natural consistency property with respect to the ancestral margin of
the DAG.

A fundamental challenge in working with counterfactual quantities such
as counterfactual explainability is that they are generally only
partially identified. This is because we can only identify
distributions of ``single-world'' potential outcomes using empirical data,
but counterfactual quantities depend on the distribution of ``cross-world''
potential outcomes. We illustrate this problem with an example in
\Cref{sec:identification-1}. If the basic potential outcomes are
comonotone, the counterfactual explainability is point
identifiable. Under this assumption, we discuss some estimation
methods based on Monte-Carlo in \Cref{sec:estimation-1}.

In \Cref{sec:simulation}, we apply the notion of counterfactual
explainability to a real dataset to explain income inequality
by sex, race, and educational attainment. Our analysis shows the
explainability of sex and educational attainment is substantial and
exhibits an interesting pattern in relation to age. We conclude the
article with some further discussion in
\Cref{sec:discussion}. Technical proofs of the Theorems in this
article can be found in \Cref{app:proofs}.

\noindent\textbf{Conventions}.
We use $[K]$ to denote the set $\{1, 2, \ldots, K\}$.
For $\calS \subseteq [K]$, we use $|\calS|$ to denote its cardinality and $-\calS$ to denote its complement in $[K]$.
We sometimes write a singleton set $\{k\}$ as $k$.
For $w \in \RR^K$, we use $w_{\calS}$ to denote the sub-vector
$(w_k)_{k \in \calS}$.
For $w$, $w' \in \RR^K$, we introduce the hybrid point $(w_{\calS}, w'_{-\calS})$ where the $k$-th coordinate is $w_k$ for $k \in \calS$ and $w'_k$ for $k \notin \calS$.

\section{Functional ANOVA and global sensitivity analysis}
\label{sec:func-anova-glob}

\subsection{Literature review}\label{sec:literature-review}

We briefly review functional ANOVA
\citep{hoeffding1948} and global sensitivity analysis
\citep{sobol1993sensitivity}. Any function $f$ of (probabilistically)
independent variables $W_1,\dots, W_K$ can always be decomposed as $f(W) =
\sum_{\calS \subseteq [K]} f_{\calS}(W_{\calS})$, where the terms can
be obtained inductively via (let $f_\emptyset(W) = \E[f(W)]$)
\[
  f_{\calS}(w_{\calS}) = \E\Big[ f(W) - \sum_{\calS' \subset \calS}
  f_{\calS'}(W) \mid W_{\calS} = w_{\calS}\Big],\quad \text{for}~\calS
  \subseteq [K],
\]
and they are orthogonal in the sense that $\E[f_{\calS}(W_{\calS}) f_{\calS'}(W_{\calS'})]=0$ for all different $\calS, \calS' \subseteq [K]$. This is often referred to as \emph{Hoeffding's decomposition} and implies that the total variance of $f(W)$ can be decomposed as
\[
  \var(f(W)) = \sum_{\calS \subseteq [K]} \sigma_{\calS}^2,\quad \text{where}~\sigma_{\calS}^2 := \var(f_{\calS}(W_{\calS})).
\]
The last equation is often referred to as \emph{functional ANOVA}.

The functional ANOVA decomposition provides the basis for a variety of variable
importance measures that are useful to assess the ``global sensitivity'' of model output to the input variables (``global'' because they do not just investigate local perturbations to $W$). For a subset $\calS \subseteq [K]$, \emph{Sobol's lower
  and upper sensitivity indices} are defined, respectively, as
\[
  \underline{\tau}^2_{\calS} 
  = \sum_{\calS' \subseteq \calS} \sigma^2_{\calS'}\quad \text{and}
  \quad \overline{\tau}^2_{\calS}
  = \sum_{\calS' \cap \calS \neq \emptyset} \sigma^2_{\calS'}.
\]
This is analogous to the definition of belief and plausibility
functions in the Dempster-Shafer theory
\citep{dempsterUpperLowerProbabilities1967,shafer1976mathematical}.
The \emph{superset importance} \citep{hooker2004discovering} is defined as
\begin{align*}
  \overline{\sigma}^2_\calS = \sum_{\calS' \supseteq \calS} \sigma^2_{\calS'}.
\end{align*}
These indices have applications in defining the effective dimension
and approximating a function
\citep{owenDimensionDistributionQuadrature2003,liu2006estimating,hartApproximationTheoreticPerspective2018}.
\emph{Shapley's value} is an axiomatic way of attribution in cooperative games \citep{shapley1953value}. If we use Sobol's lower index as the value of any subset $\calS$, the Shapley value of $W_k$ is given by \citep{owen2014sobol}
\begin{align*}
  \phi_k = \frac{1}{K} \sum_{\calS \subseteq [K]\setminus\{k\}} \left(\binom{K - 1}{|\calS|} \right)^{-1} (\underline{\tau}^2_{\calS \cup \{k\}} - \underline{\tau}^2_{\calS}) = \sum_{\{k\} \subseteq \cal S \subseteq [K]} \sigma_\calS^2/|\calS|.
\end{align*}
The variance of any interaction term is distributed equally to the factors involved, and various extensions of Shapley's value to incorporate non-zero interaction terms have been proposed \citep{grabisch1999axiomatic, rabitti2019shapley, sundararajan2020shapley}.
One major limitation of these measures is that functional analysis of
variance requires independent input variables. Although there are
extensions to the dependent case \citep[see
e.g.][]{hooker2007generalized,chastaing2012generalized,owen2017shapley},
these variable importance measures do not share all the good properties of the corresponding measures in the
independent case.


It is not necessary to obtain the full ANOVA decomposition to compute
Sobol's indices and superset importance. In fact, we have
\begin{equation}
  \label{eq:pick-freeze}
  \underline{\tau}^2_{\calS} = \Cov(f(W), f(W_{\calS}, W'_{-\calS}))
  \quad \text{and} \quad
  \overline{\tau}_{\calS}^2 = \frac{1}{2} \E[ \{f(W) - f(W_{\calS}',
  W_{-\calS})\}^2],
\end{equation}
where $W'$ is an independent and identically distributed copy of $W$.
The first identity is due to \citet{sobol1993sensitivity} and the
second is due to \citet{jansenAnalysisVarianceDesigns1999}.
This is known as the \emph{pick-freeze} method in the global
sensitivity analysis and quasi-Monte Carlo literature
\citep{sobol2001global}.
For the superset importance, we can also
write it as the mean square of a contrast. To this end, we define the
interaction contrast of a function $f$ with respect to a non-empty
$\calS \subseteq [K]$ anchored at $w$ and evaluated at $w'$ as
\begin{equation}
  \label{eq:interaction-contrast}
  I_{\calS}(w, w') = \sum_{\calS' \subseteq \calS} (-1)^{|\calS| - |\calS'|} f(w'_{\calS'}, w_{-\calS'}).
\end{equation}
By convention, $I_{\emptyset}(w, w') = f(w)$ is the value of $Y$
evaluated at the anchor $w$. The \emph{anchored decomposition} of $f$
\citep{sobol1969multidimensional,kuo2010decompositions} refers to
the identity
\[
  f(w') = \sum_{\calS \subseteq [K]} I_{\calS}(w, w'),~\text{for all}~w~\text{and}~w',
\]
which is an instance of the inclusion-exclusion principle (also called
Möbius inversion). Note that this decomposition does not
require $W_1,\dots,W_K$ to be variationally
independent. It is shown in \citet[Theorem 1]{liu2006estimating} that
\begin{equation}
  \label{eq:superset-pick-freeze}
  \overline{\sigma}^2_{\calS} = 2^{-|\calS|} \var\{I_{\calS} (W,
  W')\},\quad \text{for all}~\calS \subseteq [K].
\end{equation}
By \eqref{eq:pick-freeze}, we have $\overline{\sigma}^2_{\calS} =
\overline{\tau}^2_{\calS}$ when $\calS$ is a
singleton. \cite{owenVarianceComponentsGeneralized2013} further
considered accelerated computation for generalized Sobol's indicies
which are linear combinations of all possible second-order moments of
$f(W)$ based on fixing two subsets of input variables.

\begin{example}
When $K = 2$, the anchored decomposition is given by
\[
  f(w') = I_{\emptyset}(w, w') + I_{\{1\}}(w, w') + I_{\{2\}}(w, w') + I_{\{1,2\}}(w, w'),
\]
where
\begin{align*} 
  &I_{\{1\}}(w, w') = f(w_1',w_2) - f(w_1, w_2), \quad I_{\{2\}}(w, w') = f(w_1,w_2') - f(w_1, w_2) \\
  &I_{\{1,2\}}(w, w') = f(w_1' , w_2') - f(w'_1, w_2) - f(w_1, w'_2) + f(w_1, w_2).
\end{align*}
By \eqref{eq:pick-freeze} and \eqref{eq:superset-pick-freeze},
\begin{align*}
  \overline{\tau}_{\{1\}}^2 = \Var(I_{\{1\}}(W, W')) / 2 = \overline{\sigma}_{\{1\}}^2
                                 \quad
                                \text{and} \quad
  \overline{\sigma}_{\{1,2\}}^2 = \Var(I_{\{1,2\}}(W, W')) / 4.
\end{align*}
Intuitively, if $w$ and $w'$ are close, $I_{\{1,2\}}(w,w')$ is a finite difference
approximation to the partial derivative $\partial^2 f /(\partial w_1
\partial w_2)$ at $w$. This is also commonly used to measure
interaction effect in the causal inference literature
\citep{vanderweele2014attributing,vanderweele2015explanation},
especially in factorial experiments \citep{dasgupta2015causal}.
\end{example}

\subsection{Unification of variable importance measures}
\label{sec:unif-vari-import}

We next show that various notions of subset variable importance in
\Cref{sec:literature-review} can be unified using what we call the
explanation algebra generated by the variables $W_1,\dots,W_K$.

\begin{definition}[Explanation algebra]\label{defi:explainability.lattice}
  The \emph{explanation algebra} $\mathcal{E}(W)$ generated by a
  collection of variables $W = (W_1,\dots,W_K)$ is the Boolean algebra
  defined by the conjunction $\vee$, disjunction $\wedge$, and
  negation $\neg$ that is isomorphic to the set algebra for the power
  set of $\{0,1\}^{K}$ (each element in this set algebra is a
  collection of binary vectors of length $K$) defined by set union
  $\cup$, set intersection $\cap$, and set complement under the map
  $W_k \mapsto \{w \in \{0,1\}^{K}: w_k = 1\}$, $k \in [K]$.
\end{definition}

\begin{table}
  \caption{Interpretation of the explanation algebra ($K = 2$).}
  \label{tab:ew} \centering
  \begin{tabular}[t]{lll}
    \toprule
    Clause in $\mathcal{E}(W)$ & Maps to & Interpretation \\
    \midrule
    $W_1$ & $\{(1,0), (1,1)\}$ & Explained by $W_1$ \\
    $W_2$ & $\{(0,1), (1,1)\}$ & Explained by $W_2$ \\
    $\neg W_1$ & $\{(0,0), (0,1)\}$ & Not explained by $W_1$ \\
    $\neg W_2$ & $\{(0,0), (1,0)\}$ & Not explained by $W_2$ \\
    $W_1 \wedge W_2$ & $\{(1,1)\}$ & Explained by the interaction of
                                     $W_1$ and $W_2$ \\
    $W_1 \vee W_2$ & $\{(1,0), (0,1), (1,1)\}$ & Explained by $W_1$ and
                                                 $W_2$ \\
    $\neg (W_1 \vee W_2)$ & $\{(0,0)\}$ & Not explained by $W_1$ or
                                          $W_2$ \\
    $W_1 \wedge (\neg W_2)$ & $\{(1,0)\}$ & Explained solely by $W_1$ \\
    $W_2 \wedge (\neg W_1)$ & $\{(0,1)\}$ & Explained solely by $W_2$ \\
    \bottomrule
  \end{tabular}
\end{table}

\Cref{tab:ew} demonstrates the interpretation of some clauses in
$\mathcal{E}(W)$ when $K = 2$. When $C_1, C_2 \in \mathcal{E}(W)$ are
mapped to disjoint subsets of $\{0,1\}^K$ (so $C_1 \wedge C_2$ is
mapped to the empty set), we write $C_1 \wedge C_2 = \emptyset$; for
example, $W_1 \wedge (\neg W_1) = \emptyset$. When using the
explanation algebra $\mathcal{E}(W)$ below, we will treat $W$ as an
unordered sequence of variables. For example, both $\mathcal{E}(W_1,W_2,W_3)$ and
$\mathcal{E}(W_3,W_2,W_1)$ contain $W_1 \vee W_2$, but we treat them as the same clause; see also the Symmetry property in \Cref{defi:axiom} below.


\begin{theorem} \label{thm:xi-equivalent}
  Let $W_1,\dots,W_K$ be independent and $f$ be a given function of $W
  = (W_1,\dots,W_K)$ such that $\Var(f(W)) < \infty$. Let $\xi_1$ be any
  probability measure on $\mathcal{E}(W)$ such that
  \begin{equation}
    \label{eq:xi-anova}
    \xi_1\left((\wedge_{k \in \calS} W_k) \wedge (\wedge_{k \notin \calS} \neg W_k)\right) = \frac{\sigma_{\calS}^2}{\var(f(W))},\quad\text{for all}~\calS \subseteq [K].
  \end{equation}
  Let $\xi_2$ be any probability measure on $\mathcal{E}(W)$ such that
  \begin{equation}
    \label{eq:xi-sobol-lower}
    \xi_2\left(\neg (\vee_{k \not \in \calS} W_k)\right) =
    \frac{\underline{\tau}^2_{\calS}}{\var(f(W))},\quad\text{for
      all}~\calS \subseteq [K].
  \end{equation}
  Let $\xi_3$ be any probability measure on $\mathcal{E}(W)$ such that
  \begin{equation}
    \label{eq:xi-sobol-upper}
    \xi_3(\vee_{k \in \calS} W_k) =
    \frac{\overline{\tau}^2_{\calS}}{\var(f(W))},\quad\text{for
      all}~\calS \subseteq [K].
  \end{equation}
  Let $\xi_4$ by any probability measure on $\mathcal{E}(W)$ such
  that
  \begin{equation}
    \label{eq:xi-superset}
    \xi_4(\wedge_{k \in \calS} W_k) =
    \frac{\overline{\sigma}^2_{\calS}}{\var(f(W))},\quad\text{for
      all}~\calS \subseteq [K].
  \end{equation}
  Then $\xi_1 = \xi_2 = \xi_3 = \xi_4$.
\end{theorem}

Let $\xi$ denote any of the four equivalent probability measures in
\Cref{thm:xi-equivalent}. It is not the first time that this
probability measure has been considered:
\citet{caflischValuationMortgagebackedSecurities1997} and
\citet{liu2006estimating} have used the ``dimension distribution''
---generating a random set $\calS \subseteq [K]$ with
probability in proporition to $\sigma_{\calS}^2$ (essentially
the probability measure $\xi_1$)---to define the ``effective dimension'' and
``mean dimension'' of the function $f$. The existence and uniqueness
of $\xi_1$ (and thus $\xi$)
is guaranteed by the functional ANOVA decomposition (in particular,
the fact that $\sigma_{\calS}^2 \geq 0$). In
\Cref{sec:inclusion-exclusion}, we provide an alternative proof of the
existence of $\xi_3$ using the anchored decomposition. A key lemma in
our proof rewrites superset importance as
an covariance between two interaction constrats:
\begin{align*}
  \overline{\sigma}^2_{\calS \cup \calS'} = \left(-1\right)^{|\calS| + |\calS'|}
  \cov\left(I_{\calS}(W, W'), I_{\calS'}(W, W')\right),\quad \text{for
  all disjoint}~\calS,\calS' \subset [K].
\end{align*}
This provides an alternative and perhaps more intuitive way to
express the computational formula in \citet[Theorem
5.2]{owenVarianceComponentsGeneralized2013} for superset
importance.

\Cref{thm:xi-equivalent} indicates a change of
perspective: instead of thinking
about the ANOVA decomposition, Sobol's lower and upper indices, and the
superset importance as different notions of variable importance for
the same subset $\calS$, we can view them as measures of different
explanation clauses in a single underlying probability measure $\xi$. This allows us to, for example,
visualize explanability using Venn's diagram
(see \Cref{ex:xi-simple-functions} below). Moreover, we immediately have the
following useful properties:
\begin{enumerate}
\item $\xi(\vee_{k \in \calS} W_k) = 1$.
\item $\xi(\vee_{k \in \calS'} W_{k}) \geq  \xi(\vee_{k
    \in \calS} W_{k})$ and $\xi(\wedge_{k \in \calS'} W_{k})
  \leq  \xi(\wedge_{k \in \calS} W_{k})$ for any $ \calS
  \subseteq \calS' \subseteq [K]$.
\item $\xi(W_1) + \dots \xi(W_K) \geq \xi(\vee_{k \in [K]}
  W_k)$.
\end{enumerate}
The second property shows that if the interaction explainability of
$\calS$ is zero, so is any of its superset (which represents
higher-order interactions). This hierarchy may be useful when trying
to construct tests for vanishing interactions
\citep{hooker2004discovering}. The third property is
basically the Efron-Stein inequality \citep{efron-stein}:
\[
  \Var(f(W)) \leq \sum_{k=1}^K \Var(f(W) - f(W'_k, W_{-k})) = \sum_{k=1}^K \Var(I_{\{k\}}(W, W')).
\]
By using the inclusion-exclusion principle or equivalently the
interlacing Boole-Bonferroni inequalities, one
can further tighten this upper bound and provide lower bounds for
$\Var(f(W))$ using high-order interaction contrasts.

We demonstrate the explainability measure $\xi$ with some simple multivariate functions.

\begin{figure}[t]
  \centering
  \begin{minipage}{0.24\textwidth}
    \centering
    \includegraphics[clip, trim = 0.5cm 0cm 0.5cm 0cm, width = 1.1\textwidth]{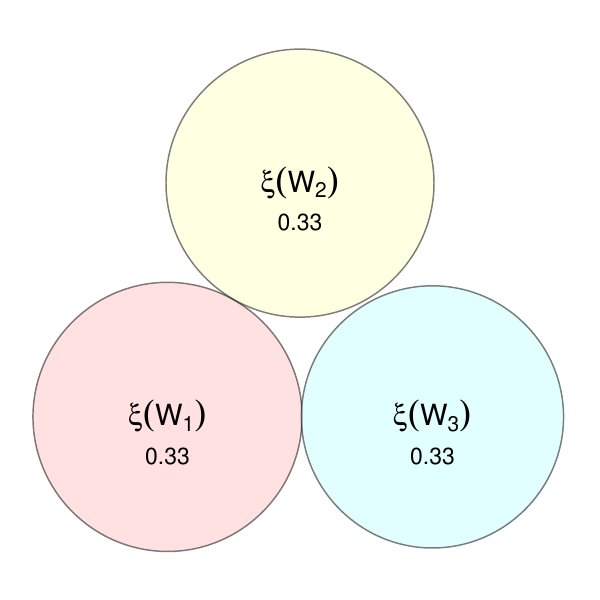}
    \caption*{(a) Linear.}
  \end{minipage}
  \begin{minipage}{0.24\textwidth}
    \centering
    \includegraphics[clip, trim = 1cm 0cm 1cm 0cm, width = 1\textwidth]{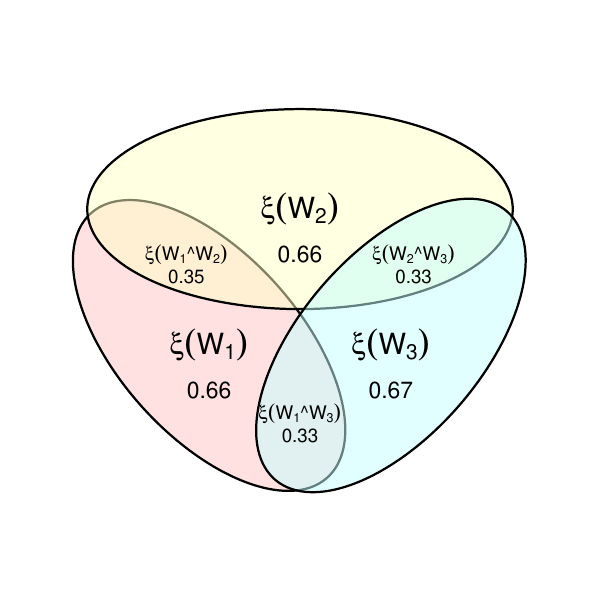}
    \caption*{(b) Quadratic.}
  \end{minipage}
  \begin{minipage}{0.24\textwidth}
    \centering
    \includegraphics[clip, trim = 1cm 0cm 1cm 0cm, width = 1\textwidth]{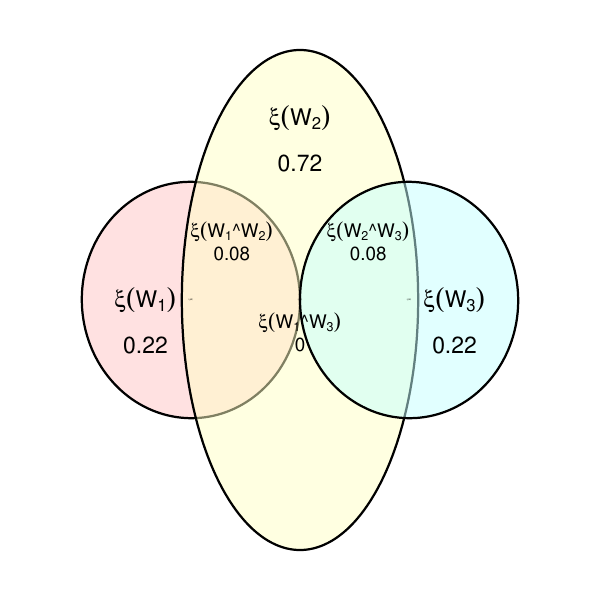}
    \caption*{(c) Single-layer NN.}
  \end{minipage}
  \begin{minipage}{0.24\textwidth}
    \centering
    \includegraphics[clip, trim = 1cm 0cm 1cm 0cm, width = 1\textwidth]{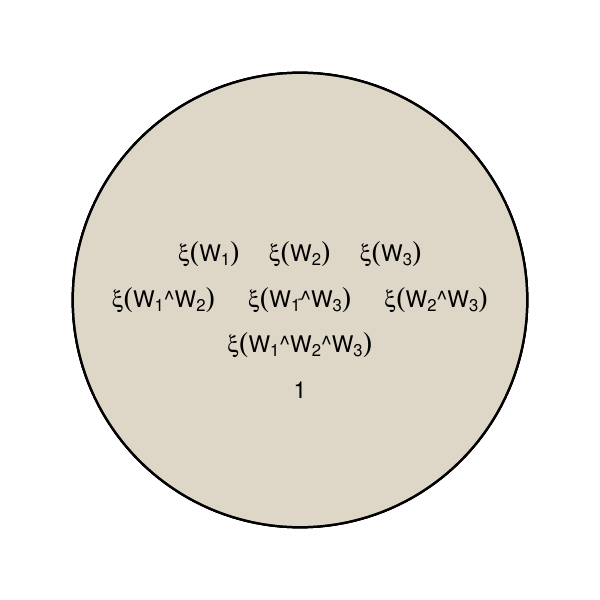}
    \caption*{(d) Multilinear.}
  \end{minipage}
  \caption{Explainability of causally independent factors. Each value is approximated using Monte Carlo with $10^6$ random draws.}
  \label{fig:simulation.independent}
\end{figure}

\begin{example} \label{ex:xi-simple-functions}
  \Cref{fig:simulation.independent} shows the the value of $\xi$ in a Venn diagram for the following functions of three variables: (1) linear: $ f(w) = \sum_{k=1}^3 w_k $; (2) quadratic polynomial: $ f(w) = w_1 w_2 + w_1 w_3 + w_2 w_3 $; (3) single-layer neural network (sigmoid activation): $ f(w) = (1 + e^{10w_1 + 10 w_2})^{-1} + (1 + e^{10 w_2 + 10 w_3})^{-1}$; (4) multilinear monomial: $ f(w) = w_1 w_2 w_3 $. 

\end{example}



\subsection{Axiomatization of total explainability}
\label{sec:axiom-total-expl}


Next, we provide an axiomatic justification of Sobol's upper
sensitivity index, which in turn provides a justification of the
explainability measure $\xi_3$ and any of its equivalent definition in
\Cref{thm:xi-equivalent}.

Our argument is based on the characterization of the variance functional
in \citet{mattner1999cumulants}. To this end, we will need to introduce some notation. Let the probability space $\P$ be
given. Let $L_r(\P)$ denote the collection of all real-valued
random variables defined on this probability space with finite $r$-th
moment, $L_r(\P) = \{X: \E(|X|^r) <
\infty \}$, $r < \infty$, and let $L_{\infty}(\P) = \cap_{r=1}^{\infty}
L_r(\P)$ collects variables with all finite moments. For $r =
1,2,\dots,\infty$, let $L_r^K(\P)$ denote the set of $K$-dimensional
random vectors whose components belong to $L_r(\P)$ and define
\[
  L_{r,\indep}^K(\P) = \left\{ W = (W_1, \dots, W_K) \in L_r^K : W_1, \dots, W_K
    \text{ are independent} \right\}.
\]
It is well known since \citet{fisher1918correlation} (who coined the
term \emph{variance}) that the variance is \emph{additive} in the
sense that
\[
  \Var(W_1 + W_2) = \Var(W_1) + \Var(W_2),~\text{for all}~(W_1, W_2) \in L_{2,\indep}(\P).
\]
\citet[1.14]{mattner1999cumulants} showed that, up to a
multiplicative constant, $\Var$ is in fact the \emph{only functional}
on $L_{\infty}(\P)$ that is additive, continuous, and
non-negative.\footnote{We follow
  the definitions in \cite{mattner1999cumulants} regarding topology and
  continuity. Specifically, we equip each space \(L_r\) with a
  topology induced by the weighted total variation metric \(d_r\),
  defined as
  \[
    d_r(\P, \Q) := \int (1 + |x|^r) \, \diff |\P - \Q|(x), \quad \text{for } \P, \Q\in L_r.
  \]
  (Because $\Var$ only depends on the distribution of the random
  variable, we can regard two variables with the same probability distribution
  as in the same equivalence class. Additivity then means that $\Var(\P
  \ast \Q) = \Var(\P) + \Var(\Q)$ where $\ast$ is convonlution of
  distributions.)
  We then topologize \(L_\infty\) using the family of metrics \((d_r : r
  \in \mathbb{N}_+)\).
}



Let $L_r(W) = \{Y \in L_r(\P): Y = f(W)~\P\text{-almost surely}\}$
collect all $L_r$ variables that are measurable functions of
$W$. We will write $\zeta_W(C \Rightarrow Y)$ as the explainability of
the clause $C \in \mathcal{E}(W)$ to $Y \in L_r(W)$ given the
explanatory variables $W$. We will show that $\xi$ in \Cref{thm:xi-equivalent} is the unique (normalized) explanability measure that satisfies the properties in the following Definition.

\begin{definition}\label{defi:axiom}
  Let $r \geq 2$ be given. We say the collection of functions
  \[\unnormalxi_{W}(\cdot
    \Rightarrow \cdot): \mathcal{E}(W) \times L_r(W) \to
    \mathbb{R}, \quad \text{for}~W \in L_{r,\indep}^K(\P),~K \geq 1,
  \]
  define an \emph{explainability measure} if they satisfy the
  following properties:
  \begin{enumerate}
  \item {\bf Symmetry:} for any $W \in L_{r,\indep}^K(\P)$, $C \in \mathcal{E}(W)$, $Y \in
    L_r(W)$, and permutation $\pi(W)$ of $W$ (e.g., $\pi(W_1,W_2) =
    (W_2, W_1)$), we have
    $
    \unnormalxi_{W}(C \Rightarrow Y) = \unnormalxi_{\pi(W)}(C
    \Rightarrow Y).
    $
  \item {\bf Totality:} for all $W \in L_{r,\indep}^K(\P)$ and $Y \in L_r(W)$, we
    have
    \[
      \zeta_{W}(\vee_{k=1}^K W_k \Rightarrow Y) =
      \zeta_Y(Y \Rightarrow Y) \geq 0.
    \]
  \item {\bf Continuity:} $\zeta_Y(Y \Rightarrow Y)$ is a continuous
    functional of the distribution of $Y \in L_r(\P)$.
  \item {\bf Linearity}: for any $K \geq 2$, $W \in
    L_{r,\indep}^K(\P)$, and $Y = f(W) \in L_r(W)$, we have
    \[
      \unnormalxi_W\left(\vee_{k=2}^K W_k \Rightarrow Y\right) = \int
      \unnormalxi_{W_{-1}}\left(\vee_{k = 2}^K W_k \Rightarrow
        Y(w_1)\right) \diff\P(w_1),
    \]
    where $Y(w_1) = f(w_1, W_{-1})$ and $\P(w_1)$ is the marginal
    distribution of $W_1$.\footnote{By disintegrating the
      joint probability measure of $(W_1, W_2)$, it is easy to see that
      the functional representation of $Y$ does not matter in this
      definition. That is, if $f(W) = g(W)$ with probability $1$, then
      $\P(f(w_1,W_{-1}) = g(w_1, W_{-1})) = 1$ for almost all $w_1$.}
  \item {\bf Additivity I}: for all $W \in L_{r,\indep}^2(\P)$, we have
      \[
        \zeta_{W}(W_1 \vee W_2 \Rightarrow W_1 + W_2) =
        \zeta_{W}(W_1 \vee W_2 \Rightarrow W_1) + \zeta_{W}(W_1 \vee W_2 \Rightarrow W_2).
      \]
  \item {\bf Additivity II}: for all $W \in L_{r,\indep}^K(\P)$, $Y \in L_r(W)$,
    and $C_1,C_2 \in \mathcal{E}(W)$ such that $C_1 \wedge C_2 =
    \emptyset$, we have
    \[
      \unnormalxi_W(C_1 \vee C_2 \Rightarrow Y) = \unnormalxi_W(C_1
      \Rightarrow Y) + \unnormalxi_{W}(C_2 \Rightarrow Y).
    \]
  \end{enumerate}
\end{definition}



\begin{theorem} \label{thm:uniqueness}
  Let $r \geq 2$ be given. Suppose $\zeta$ satisfies axioms 1-5 in
  Definition \ref{defi:axiom}. Then there exists a constant $c > 0$
  such that
  \begin{equation*}
    \unnormalxi_W(\vee_{k \in \calS} W_k \Rightarrow Y) = c \E[\var(f(W)
    \mid W_{-\calS})] = c \overline{\tau}^2_{\calS},\quad \text{for
      all}~\calS \subseteq [K].
  \end{equation*}
  Moreover, if $\zeta$ additionally satisfies axiom 6 in
  \Cref{defi:axiom} (so $\zeta$ is an explanation measure), then
  \[
    \frac{\zeta_W(C \Rightarrow Y)}{\zeta_W(\vee_{k=1}^K W_k
      \Rightarrow Y)} = \xi(C),\quad\text{for all}~W \in
    L_{r,\indep}^K(\P),~Y \in L_r(W),
  \]
  where $\xi$ is any of the probability measure defined in
  \Cref{thm:xi-equivalent}.
\end{theorem}



As shown in the proof of this Theorem, the Symmetry,
Non-negativity, Totality, Continuity, Linearity, and Additivity I
properties ensure that $\zeta_Y(Y \Rightarrow Y) = \Var(Y)$ using the
result in \citet{mattner1999cumulants}. Additivity II is simply the
finite-additivity of probability measures and allows us to extend
Sobol's upper index or the notion of total explainability to the whole
explanation algebra $\mathcal{E}(W)$. With these in mind, the key
axiom in \Cref{defi:axiom} that allow us to justify Sobol's upper
index is Linearity. Heuristically, the Linearity axiom says that the
total explainability of $W_2,\dots,W_K$ to $Y$ is an average of that
explainability to the potential outcome $Y(w_1)$.

\citet{hartApproximationTheoreticPerspective2018} gives an approximate
theoretic perspective on $\overline{\tau}_{\calS}^2$ with dependent
$W_1,\dots,W_K$ based on the observation that the conditional
expectation $\E[f(W) \mid W_{-\calS}]$ is the best least-squares
approximation to $f(W)$ among all functions of $W_{-\calS}$. Reviewing
the proof of \Cref{thm:uniqueness}, this choice can be justified in a
similar axiomatic way by requiring the Linearity
property to hold for all $W \in L_r^K(\P)$ with
$Y(w_1)$ replaced by $\E[Y \mid W_{-1}]$. It is shown in
\citet{hartApproximationTheoreticPerspective2018} that this total
index corresponds to defining the individual index as suggested by
\citet{liGlobalSensitivityAnalysis2010}:
\[
  \sigma_{\calS}^2 = \frac{\Cov(f_{\calS}(W_{\calS}),  f(W))}{\var(f(W))},
  \quad \calS \subseteq [K].
\]
Even though these indices still sum up to $1$, it is not guaranteed
that they are non-negative. In other words, the explainability measure
$\xi$ defined this way is generally a signed measure with total $1$. 

\section{Counterfactual explainability}\label{sec:independent}

\subsection{Motivation: Inconsistency of non-causal explanations}
\label{sec:incons-non-caus}

The functional ANOVA-based approach to global sensitivity analysis in
\Cref{sec:func-anova-glob} has several limitations:
\begin{enumerate}
\item The explanatory variables $W_1,\dots,W_K$ are required to be
  independent. (This can be relaxed using the proposal in
  \citet{liGlobalSensitivityAnalysis2010} and
  \citet{hartApproximationTheoreticPerspective2018} as discussed above,
  but $\xi$ loses the nonnegative property.)
\item The outcome variable $Y$ is required to be a deterministic
  function of $W$.
\item The explainability measure heavily depends on which explanatory
  variables are included.
\end{enumerate}

We demonstrate the last point about inconsistency in the example below.

\begin{example}[Inconsistency of global sensitivity analysis]\label{exam:Sobol.inconsistent}
  Let $\xi_W(\cdot \Rightarrow Y)$ denote the explainability measure in
  \Cref{thm:xi-equivalent}. Let $W_1, W_2, W_3$ be Rademacher random
  variables such that, for $(w_1,w_2,w_3) \in \{-1,1\}^3$,
  \[
    \P(W_1 = w_1, W_2 = w_2, W_3 = w_3) =
    \begin{cases}
      1/4, & \text{if}~w_1 w_2 w_3 = 1, \\
      0, & \text{otherwise}.
    \end{cases}
  \]
  It is easy to verify that $W_1, W_2, W_3$ are pairwise independent.
  Consider the following functions:
  \[
    f_1(w_1, w_2) = w_2 \quad \text{and} \quad f_2(w_1, w_3) = w_1 w_3.
  \]
  Let $Y = f_1(W_1,W_2) = f_2(W_1, W_3)$ (the second equality holds
  with probability $1$). However, we have
  \[
    \xi_{(W_1,W_2)}(W_1 \Rightarrow Y) = 0 \quad \text{and} \quad
    \xi_{(W_1,W_3)}(W_1 \Rightarrow Y) = 1.
  \]
\end{example}

There are two ways to explain the paradoxical observation in
\Cref{exam:Sobol.inconsistent}. The first explanation is that the
explainability of $W_1$ to $Y$ depends on the context (i.e., what other
explanatory variables are included). The second explanation is that
even though $Y_1 = f_1(W_1,W_2)$ and $Y_2 = f_2(W_1, W_3)$ are equal
with proability $1$, they represent different objects: $Y_1$
represents the causal model
specified by $f_1(w_1,w_2) = w_2$, and $Y_2$ represents the causal
model specified by $f_2(w_1,w_3) = w_1 w_3$. Although there is nothing
inherently wrong in the first explanation to \Cref{exam:Sobol.inconsistent},
we will see that the second approach motivates a general notion of counterfactual explainability that not only addresses the inconsistency problem but also the other limitations of functional ANOVA.

\subsection{Graphical causal model and counterfactual
  explainability}\label{sec:dependent}


Our definition of counterfactual explainability requires a DAG
causal model for the explanatory and outcome variables
\citep{pearl2009,hernanCausalInferenceWhat2023}. We first briefly
define the causal model. Let $G$ be a directed
acyclic graph (DAG) with vertices being some random variables
$V_1,\dots,V_K$; by acyclic, we mean the graph does not contain a
cycle like $V_k \rightarrow \cdots \rightarrow V_k$. We use $\pa(k)$
to denote the set of vertices with a
directed edge pointing to $V_k$ (aka ``parents'' of $V_k$) in
$G$. We use $V_k(v_\calS)$ to denote the \emph{potential outcome} of
$V_k$ under an intervention that sets $V_\calS$ to $v_\calS$, and
define the potential outcomes schedule as the collection of all such
variables (which we assume are defined in the same probability space):
\begin{align*}
  V(\cdot) = \left(V_k(v_\calS): k \in [K], \calS \subseteq [K],
  v_\calS~\text{in the support of}~V_\calS \right).
\end{align*}
As a convention, we use $V_k$ to also denote the potential outcome of
$V_k$ when there is no intervention ($\calS = \emptyset$).
The schedule of all basic potential outcomes of $V_k$ is defined as
\[
  V_k^* = (V_k(v_{\pa(k)}): v_{\pa(k)}~\text{in the support of $V_{\pa(k)}$}).
\]
For any $\calS \subseteq [K]$, let $V_\calS(\cdot)$ and $V_\calS^*$
collects all the potential outcomes and basic potential outcomes for
$V_\calS$, respectively.

\begin{definition}[Causal Markov model] \label{def:causal-markov}
  We say a probability distribution $\P$ on $V(\cdot)$ is \emph{causal
    Markov} with respect to a DAG $\calG$ with vertex set $W$ if the
  following are true:
  \begin{enumerate}
  \item The next event has probability $1$:
    \begin{equation}
      \label{eq:basic-consistency}
      V_k(v_{\calS}) = V_k(v_{\pa(k) \cap \calS}, V_{\pa(k) \setminus
        \calS}(v_{\calS})),~\text{for all}~k \in [K], ~\calS \subseteq
      [K], ~\text{and}~v_{\calS}.
    \end{equation}
  \item The basic potential outcomes schedules $V_1^*, \dots, V_K^*$
    are independent.
  \end{enumerate}
\end{definition}

The first assumption in the above Definition is often referred to as
recursive substitution or consistency property of the potential
outcomes in the literature. The second assumption in Definition
\ref{def:causal-markov}
asserts that when we collect the basic potential outcomes for
different variables, they are independent. This definition is
equivalent to the nonparametric structural equation model with
independent errors (NPSEM-IE) in \citet{pearl2009} (see Example \ref{exam:NPSEM} below). Properties of this causal Markov model are discussed in
\citet{richardson2013single} and
\citet{zhaoStatisticalCausalModels2025}.

Note that by recursively
applying \eqref{eq:basic-consistency}, all potential outcomes are
deterministic functions of the basic potential outcomes. Therefore, if
the vertex set is $V = \{W_1,W_2,Y\}$, then there exists a function $f$
such that
\begin{equation}
  \label{eq:y-function-basic-po}
  Y = f(W_1^{*},\dots,W_K^{*},Y^{*}).
\end{equation}
To illustrate this, for the simple DAG with edges $W_1 \rightarrow Y$, $W_2 \rightarrow Y$ and $W_1 \rightarrow Y$, we have
\[
  Y = Y(W_1, W_2) = Y(W_1, W_2(W_1))
\]
is indeed a function of
$W_1^*$ (because $W_1(w) = W_1$ for all $w$), $W_2^*$ (because it collects $W_2(w_1)$ for all $w_1$), and $Y^{*}$ (because it collects all $Y(w_1,
w_2)$ for all $w_1,w_2$).

\begin{definition}[Counterfactual explainability] \label{def:counterfactual-xi}
  Given a caual Markov distribution $\P$ with respect to a DAG $G$
  with vertex set $V = (V_1,\dots, V_{K+1})$ where $V_1 =
  W_1,\dots,V_K = W_K, V_{K+1} = Y$, let $f$ be the function in
  \eqref{eq:y-function-basic-po}. The \emph{counterfactual
    explainability}
  $\xi_{G,\P}(\cdot \Rightarrow Y)$ is defined as the unique probability
  measure on $\mathcal{E}(V)$ such that
  \begin{equation}
    \label{eq:xi-g}
    \xi_{G,\P}(\vee_{k \in \calS} V_k \Rightarrow Y) =
    \frac{\var\left(f(V^*) - f(V^{*\prime}_{\calS},
        V^*_{-\calS})\right)}{2 \var(f(V^*))},\quad \text{for
      all}~\calS \subseteq [K+1].
  \end{equation}
\end{definition}

We will usually write this as $\xi_{G}$ and omit the dependence on
the probability distribution $\P$, which needs to be causal Markov
with respect to $G$. In defining counterfactual explainability, we
follow the approach in global sensitivity analysis but replace the
variables $W_k$ with their ``intrinsic noises'' $W_k^{*}$,
$k=1,\dots,K$, which are assumed to be independent in the causal
Markov model. The same idea is also suggested by
\citet{janzingQuantifyingIntrinsicCausal2024}, whose main motivation
is to quantity what they call ``intrinsic causal contribution''. In
contrast, our definition is a natural counterfactual extension of
global sensitivity analysis to the case with dependent explanations
and thus also inherits good properties of global sensitivity analysis.

In Definition \ref{def:counterfactual-xi}, the outcome variable
$Y$ itself is also included in forming the explanation algebra. In fact,
we have
\[
  \xi_{G}(Y \Rightarrow Y) = \xi_{G}\left(\neg (\vee_{k=1}^K W_k)
    \Rightarrow Y\right) = 1 - \xi_{G}\left(\vee_{k=1}^K W_k
    \Rightarrow Y\right),
\]
that is, the explanability of (the intrinsic noise of) $Y$ to itself
is what is not explained by $W$. The subscript $G$ of
$\xi$ is used to emphasize that counterfactual explainability depends
on not only the explanatory factors but also their causal DAG $G$; see
\Cref{sec:prop-count-expl} below for further discussion.

We demonstrate Definition \ref{def:counterfactual-xi} with some
examples. First, it is easy to see that Definition
\ref{def:counterfactual-xi} is consistent with the explainability
measure in global sensitivity analysis described in
\Cref{sec:func-anova-glob} in the following sense.

\begin{example}
  If $W_1,\dots,W_K$ are causally independent and $Y(w) =
  f(w)$ is a deterministic function of $w = (w_1,\dots,w_K)$ (the causal
  graph $G$ contains $W_1 \rightarrow Y,\dots,W_K \rightarrow Y$ only), then
  $W_k^{*} = W_k$ and the above definition reduces to
  \[
    \xi_{G}(C \Rightarrow Y) = \xi_W(C \Rightarrow Y),\quad
    \text{for all}~C \in \mathcal{E}(W),
  \]
  where $\xi_W(\cdot \Rightarrow Y)$ is the probability measure on
  $\mathcal{E}(W)$ defined in \Cref{thm:xi-equivalent}. The only
  difference is that $\xi_{G}(\cdot \Rightarrow Y)$ is defined on
  the larger algebra $\mathcal{E}(V)$, but $\xi_{G}(Y \Rightarrow Y)
  = 0$ because $Y^{*}$ is not random.
\end{example}

\begin{example}[NPSEM-IE]\label{exam:NPSEM}
  For a DAG $G$ with vertex set $V = (W_1,\dots,W_K,Y)$, causal models
  are often defined using NPSEM-IE \citep{pearl2009} :
  \[
    V_k = f_k(V_{\text{pa}(k)}, E_k),\quad k \in [K+1]
  \]
  for some functions $f_1,\dots,f_{K+1}$ and independent errors
  $E_1,\dots,E_{K+1}$. These equations are assumed to be ``structural'' in
  the sense that they continue to hold under interventions. In other
  words, the basic potential outcomes are defined as
  \[
    V_k(v_{\pa(k)}) = f_k(v_{\pa(k)}, E_k),\quad k \in [K+1].
  \]
  Because $E_1, \dots, E_{K+1}$ are independent, the basic potential
  outcomes $V_1^*, \dots, V_{K+1}^*$ are indeed independent, so NPSEM-IE is in the causal Markov model in Definition \ref{def:causal-markov}. By
  recursively using the structural equations, we can write $Y$ as $Y = f(E_1,\dots,E_{K+1})$. For example, for the simple DAG with edges $W_1 \rightarrow Y$, $W_2 \rightarrow Y$ and $W_1 \rightarrow Y$, we have
  \[
    Y = f_3(W_1, W_2, E_3) = f_3(W_1, f_2(W_1, E_2), E_3) = f_3(f_1(E_1),
    f_2(f_1(E_1), E_2), E_3).
  \]
  Let $Y = f(E_1,\dots,E_{K+1})$, then \eqref{eq:xi-g} can be
  rewritten as
  \begin{align}\label{defi:total.dependent.NPSEM}
    \xi_G(\vee_{k \in \calS} V_k \Rightarrow Y) =
    \frac{\var\left(f(E) - f(E'_\calS, E_{-\calS})\right)}{2
    \var(f(E))},
  \end{align}
  where $E'$ is an independent and identically distributed copy of
  $E$.
\end{example}

\begin{figure}[t]
  \centering
  \begin{minipage}{0.6\textwidth}
    \centering
    \resizebox{.99\textwidth}{!}{
      \begin{tikzpicture}
        \node[rectangle, draw, fill=yellow!10, minimum size=1cm] (E11) at (0, 2) {$E_{11}$};
        \node[rectangle, draw, fill=yellow!10, minimum size=1cm] (E12) at (0, 0) {$E_{12}$};
        \node[rectangle, draw, fill=yellow!10, minimum size=1cm] (E13) at (0, -2) {$E_{13}$};

        \node[circle, draw, fill=cyan!10, minimum size=1cm] (W11) at (2, 2) {$W_{11}$};
        \node[circle, draw, fill=cyan!10, minimum size=1cm] (W12) at (2, 0) {$W_{12}$};
        \node[circle, draw, fill=cyan!10, minimum size=1cm] (W13) at (2, -2) {$W_{13}$};

        \draw[->] (E11) -- (W11);
        \draw[->] (E12) -- (W12);
        \draw[->] (E13) -- (W13);


        \node[rectangle, draw = white, fill=blue!0, minimum width=1.5cm, minimum height=0.8cm] (sum1) at (5, 1) {$\sigma(W^\top \beta)$};
        \node[rectangle, draw = white, fill=blue!0, minimum width=1.5cm, minimum height=0.8cm] (sum2) at (5, -1) {$\sigma(W^\top \beta)$};

        \draw[->] (W11) -- (sum1) node[midway, above] {\small $\beta_{11}^{(1)} = 1$};
        \draw[->] (W12) -- (sum1) node[midway, above] {\small $\beta_{21}^{(1)} = 1$};
        \draw[->] (W12) -- (sum2) node[midway, above] {\small $\beta_{22}^{(1)} = -1$};
        \draw[->] (W13) -- (sum2) node[midway, above] {\small $\beta_{32}^{(1)} = -1$};


        \node[rectangle, draw = white, fill=blue!0, minimum width=1.5cm, minimum height=0.8cm] (layer1) at (2, -4) {1st layer};

        \node[rectangle, draw, fill=yellow!10, minimum size=1cm] (E21) at (7, 2.5) {$E_{21}$};
        \node[rectangle, draw, fill=yellow!10, minimum size=1cm] (E22) at (7, -2.5) {$E_{22}$};

        \node[circle, draw, fill=cyan!10, minimum size=1cm] (W21) at (7, 1) {$W_{21}$};
        \node[circle, draw, fill=cyan!10, minimum size=1cm] (W22) at (7, -1) {$W_{22}$};

        \draw[->] (sum1) -- (W21);
        \draw[->] (E21) -- (W21);

        \draw[->] (sum2) -- (W22);
        \draw[->] (E22) -- (W22);

        \node[rectangle, draw = white, fill=blue!0, minimum width=1.5cm, minimum height=0.8cm] (layer1) at (7, -4) {2nd layer};

        \node[circle, draw, fill=cyan!10, minimum size=1cm] (Y) at (12, 0) {$Y$};

        \draw[->] (W21) -- (Y) node[midway, above] {\small $\beta_{11}^{(2)} = 1$};
        \draw[->] (W22) -- (Y) node[midway, above] {\small $\beta_{21}^{(2)} = 1$};
      \end{tikzpicture}
    }
    \vspace{-0.25cm}
    \caption*{(a) Network structure.}
  \end{minipage} \\
  \begin{minipage}{0.45\textwidth}
    \centering
    \includegraphics[clip, trim = 0.5cm 0.5cm 0cm 0cm, width = 0.8\textwidth]{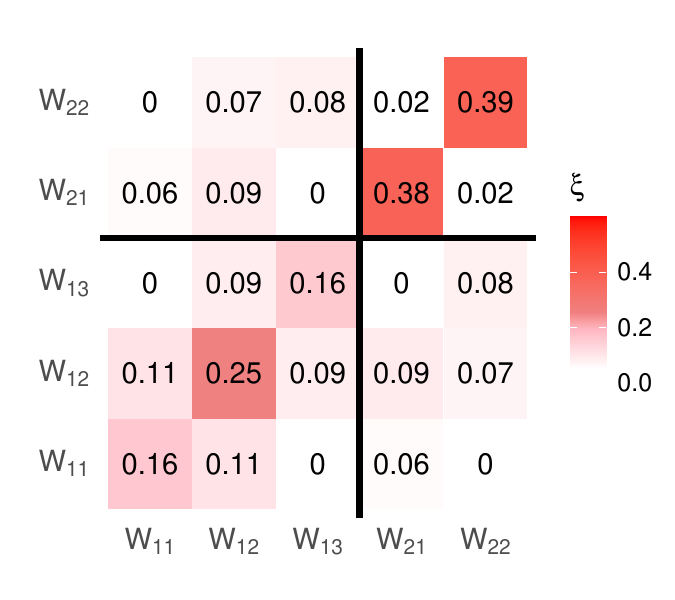}
    \caption*{(b) Equal noise variance.}
    \vspace{-0.25cm}
  \end{minipage} \quad
  \begin{minipage}{0.45\textwidth}
    \centering
    \includegraphics[clip, trim = 0.5cm 0.5cm 0cm 0cm, width = 0.8\textwidth]{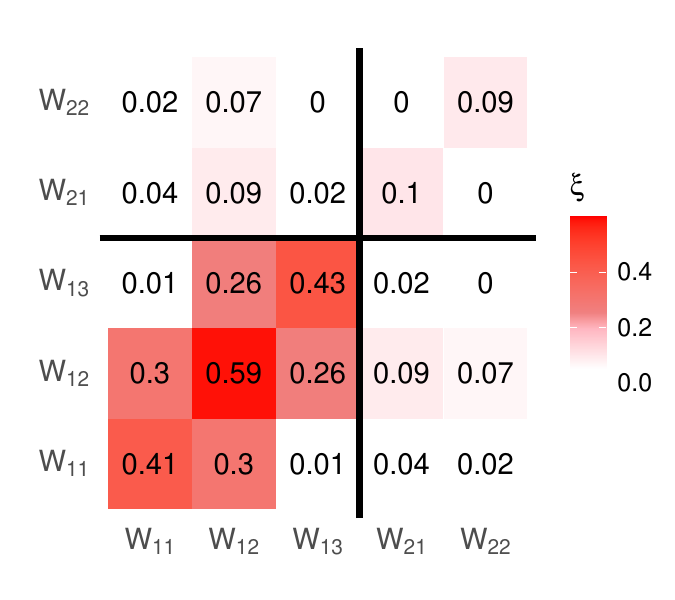}
    \caption*{(c) Second layer noise variance decreased.}
    \vspace{-0.25cm}
  \end{minipage}

  \caption{Counterfactual explainability for a two-layer neural network in which each node has under independent noise. Panel (a) shows the network structure. Panels (b) and (c) present heatmaps of counterfactual explainabilities with diagonal values representing individual input factors individually and off-diagonal values showing their pairwise interactions. In Panel (c), the standard deviations of $E_{21}$ and $E_{22}$ are decreased.} \label{fig:simulation.dependent}
\end{figure}

\begin{example}[Two-layer neural network]
  Consider the two-layer neural network with $K = 5$ nodes in \Cref{fig:simulation.dependent} (a), where the noise variables $E_{11}, E_{12}, E_{13}, E_{21}, E_{22}$ are causally independent and follow the standard normal distribution. The first-layer nodes are given by $W_{1i} = E_{1i}$ for $1 \le i \le 3$, and they are linearly combined using weights $ \beta^{(1)}$ and undergo a quadratic activation function $\sigma(x) = x^2$ to generate the second-layer nodes. Specifically, $W_{21} = \sigma(\beta^{(1)}_{11} W_{11} + \beta^{(1)}_{21} W_{12}) + E_{21}$ and similarly for $W_{22}$. The final prediction is given by $Y = \beta^{(2)}_{11} W_{21} + \beta^{(2)}_{21} W_{22}$. \Cref{fig:simulation.dependent}(b) shows the pairwise counterfactual explainability. \Cref{fig:simulation.dependent}(c) shows the counterfactual explainability for the same prediction model with $E_{21}$ and $E_{22}$ rescaled to one sixth of their original values.

\end{example}

\subsection{Ancestral consistency of counterfactual explainability}
\label{sec:prop-count-expl}

As defined above, the counterfactual explainability $\xi_G(\cdot
\Rightarrow Y)$ is a probability measure on $\mathcal{E}(V)$ and thus
retains the desirable properties of global sensitivity analysis in
\Cref{sec:func-anova-glob}. It also addresses the first two
limitations of global sensitivity analysis listed at the beginning of
\Cref{sec:independent}: $W_1,\dots,W_K$ are no longer assumed
independent, and $Y$ does not need to be a deterministic function of
$W_1,\dots,W_K$.

\begin{table}[t]
  \caption{Counterfactual explainability depends on the causal
    assumptions. In each scenario below, a NPSEM-IE with respect to
    a different DAG is given so that $W_1, W_2$ are always independent
    Rademacher variables (but they may not be causally independent).
  }\label{tab:dependent}
  \centering
  \begin{tabular}{c|c|c c c}
    \toprule
    DAG & NPSEM-IE & $\xi(W_1)$ & $\xi(W_2)$ & $\xi(W_1 \wedge
                                               W_2)$ \\ \midrule
    \begin{tikzcd}[row sep = 10pt]
      & W_2 \arrow[dr, blue] &  \\
      W_1 \arrow[rr, blue] & & Y
    \end{tikzcd}
                                & $
                                  \begin{aligned}
                                    W_1 &= E_1 \\
                                    W_2 &= E_2 \\
                                    Y &= W_1 W_2
                                  \end{aligned}
                                        $
                   & 1 & 1 & 1 \\ \hline
    \begin{tikzcd}[row sep = 10pt]
      & W_2 \arrow[dr, blue] &  \\
      W_1 \arrow[ur, blue] \arrow[rr, blue] & & Y
    \end{tikzcd}
                                & $
                                  \begin{aligned}
                                    W_1 &= E_1 \\
                                    W_2 &= W_1 E_2 \\
                                    Y &= W_1 W_2
                                  \end{aligned}
                                        $
                   & 0 & 1 & 0 \\ \hline
    \begin{tikzcd}[row sep = 10pt]
      & W_2 \arrow[dr, blue] \arrow[dl, blue] &  \\
      W_1 \arrow[rr, blue] & & Y
    \end{tikzcd}
                                & $
                                  \begin{aligned}
                                    W_2 &= E_2 \\
                                    W_1 &= W_2 E_1 \\
                                    Y &= W_1 W_2
                                  \end{aligned}
                                        $
                   & 1 & 0 & 0 \\ \bottomrule
  \end{tabular}
\end{table}

As mentioned earlier, the counterfactual explainability $\xi_{G}$
depends on the causal DAG $G$, which clarifies the presumptions for
the explanation. Thus, it is entirely possible that two agents
with different causal beliefs make different explanations; see
\Cref{tab:dependent} for an illustration. 

Next, we show that counterfactual explainability, though dependent on
the causal DAG, has certain consistency properties. We need two
graphical concepts to state the consistency property.
Let $G$ be a DAG with vertex set $V = (W_1,\dots,W_K,Y)$.
First, we say $W_{\calS} \subseteq W$ is an \emph{ancestral set} in $G$ if
$W_{\calS}$ contains all its ancestors (parents, parents of parents,
etc.) in $G$. It is easy to see that if $W_{\calS}$ is ancestral, then
\begin{equation}
  \label{eq:xi-ancestral}
  \xi_{G}(\vee_{j \in \calS} W_j \Rightarrow Y) = \frac{\Var(Y(W) -
    Y(W_{\calS}', W_{-\calS}))}{2 \Var(Y(W))},
\end{equation}
where $Y(w)$ is the potential outcome of $Y$ in an intervention that
sets $W$ to $w$. This mimics the definition of counterfactual heritability in \eqref{eq:xi-heritability} (assuming causally independent genetic and environmental factors) and the formula for Sobol's upper index in \eqref{eq:pick-freeze}. Second, we say a DAG $G'$
with vertex set $V' \supset V$ is an \emph{expansion} of $G$ if
\begin{enumerate}
\item an edge $V_j \rightarrow V_k$ is in $G$ if and only if $V_j
  \rightarrow V_k$ is also in $G'$ or there exists a directed path
  $V_j \rightarrow V_{l_1} \rightarrow \dots \rightarrow V_{l_m}
  \rightarrow V_k$ in $G'$ for some $V_{l_1},\dots,V_{l_m} \in V'
  \setminus V$.
\item for any $V_j,V_k \in V$, there exists no paths like $V_j
  \leftarrow V_{l_1} \leftarrow \dots \leftarrow V_{l_m} \rightarrow
  \dots \rightarrow V_{l_{m+n}} \rightarrow V_k$ ($m
  \geq 1$, $n \geq 0$) in $G'$ in which $V_{l_1},\dots,V_{l_{m+n}}
  \in V' \setminus V$.
\end{enumerate}
This is a special case of graph expansion for acyclic directed mixed
graphs; see \citet{zhaoStatisticalCausalModels2025}. Heuristically,
the expansion DAG $G'$ offers finer explanations about causal relations between some
variables in the original DAG $G$. See
\Cref{fig:ancestral-consistency} for an example.

\begin{theorem}[Ancestral consistency] \label{thm:ancestral-consistency}
  Let $\P$ be a causal Markov distribution with respect to a DAG $G$
  with vertex set $V = (W_1,\dots,W_K,Y)$, and $\P'$ be a causal
  Markov distribution with respect to a DAG $G'$ that is an expansion
  of $G$ with vertex set $V' = (W_1,\dots,W_{K'},Y)$ ($K' >
  K$). Suppose $\P'$ expands $\P$ in the sense that the marginal
  distribution of $\P'$ on $V(\cdot)$ agrees with $\P$. Suppose
  $W_{\calS}$ is an ancestral set in $G$, and $W_{\calS'}$ is the
  smallest ancestral set in $G'$ that contains $W_{\calS}$. Then we
  have
  \[
    \xi_{G,\P}(\vee_{j \in \calS} W_j \Rightarrow Y) =
    \xi_{G',\P'}(\vee_{j \in \calS'} W_j \Rightarrow Y).
  \]
  Moreover, if $\calS' = \calS$ (i.e., $W_{\calS}$ is also
  ancestral in $G'$), then
  \begin{equation}
    \label{eq:ancestral-consistency-2}
    \xi_{G,\P}(C \Rightarrow Y) = \xi_{G',\P'}(C \Rightarrow Y), \quad
    \text{for all}~C \in \mathcal{E}(W_{\calS}).
  \end{equation}
\end{theorem}

\begin{figure}[t] \centering
  \begin{minipage}{0.4\textwidth} \centering
    \begin{tikzcd}
      W_1 \arrow[dr, blue] & & \\
      W_2 \arrow[r, blue] \arrow[drr, blue] & W_4 \arrow[dr, blue] & \\
      W_3 \arrow[rr, blue] & & Y
    \end{tikzcd}
    \caption*{(a) Original graph $G$}
  \end{minipage}
  \begin{minipage}{0.4\textwidth} \centering
    \begin{tikzcd}
      &W_1 \arrow[dr, blue] & & \\
      &W_2 \arrow[r, blue] \arrow[drr, blue] \arrow[rd, blue] & W_4 \arrow[dr, blue]
      & \\
      W_5 \arrow[r, blue] & W_3 \arrow[r, blue] & W_6 \arrow[r, blue] & Y
    \end{tikzcd}
    \caption*{(b) Expanded graph $G'$.}
  \end{minipage}
  \caption{Demonstration of graph expansion and ancestral consistency.}
  \label{fig:ancestral-consistency}
\end{figure}

\begin{example} \label{ex:ancestral-consistency}
  Consider the DAG $G$ in \Cref{fig:ancestral-consistency}(a) and its
  expansion $G'$ in
  \Cref{fig:ancestral-consistency}(b). \Cref{thm:ancestral-consistency}
  shows that $\xi_{G}(W_3 \Rightarrow Y) = \xi_{G'}(W_3 \vee W_5
  \Rightarrow Y)$ and $\xi_{G}(C \Rightarrow Y) = \xi_{G'}(C
  \Rightarrow Y)$ for all $C \in \mathcal{E}(W_1, W_2, W_4)$.
\end{example}

It is possible to relax the ancestral assumption in
\Cref{thm:ancestral-consistency} using properties of the potential
outcomes. For example, \eqref{eq:ancestral-consistency-2} is also true
if $W_{\calS}$ has the same parents in $G$ and $G'$ because the basic
potential outcomes $W_{\calS}^{*}$ are the same in the two models. See
also \citet{janzingQuantifyingIntrinsicCausal2024} for properties of
their ``intrinsic causal contribution'' which is a more general notion
of counterfactual explainability.

\section{Identification and estimation}\label{sec:estimation}

\subsection{Identification}
\label{sec:identification-1}

Although counterfactual explainability inherits good properties of
global sensitivity analysis and has a desirable ancestral consistency
property, it cannot be fully determined even with infinite amount of
data. This ``partial identification'' problem is a central
challenge in counterfactual reasoning \citep{pearl2009,bookofwhy}. We
demonstrate it using an example.
\begin{example}
  \label{exam:identification with dependent inputs}
  Let $E_1, E_2$ be two causally independent Rademacher random variables.
  Consider two NPSEM-IE models associated with the DAG $W_1 \rightarrow W_2 \rightarrow Y$:
  \begin{align*}
    &\text{Model 1}:\quad W_1(w_1,w_2) = E_1, \quad W_2(w_1,w_2) = w_1 + w_1 E_2, \quad Y(w_1,w_2) = w_2, \\
    &\text{Model 2}:\quad W_1(w_1,w_2) = E_1, \quad W_2(w_1,w_2) = w_1 + E_2, \quad Y(w_1,w_2) = w_2.
  \end{align*}
  Under both models, the observable random variables \((W_1, W_2, Y)\) has the same joint distribution: \(W_1\) is Rademacher; given \(W_1=w_1\), $W_2$ is $w_1$ plus an independent Rademacher; and \(Y \mid (W_1=w_1, W_2=w_2)\) is \(w_2\) with probability 1.
  However, the counterfactual explainability of $W_1$ differs across the two models:
  in {Model~1},
  \[
    \xi_G(W_1 \Rightarrow Y) \;=\; \frac{\var((W_1' - W_1)(1 + E_2))}{2\,\var(Y(W))} \;=\; 1,
  \]
  whereas in {Model~2},
  \[
    \xi_G(W_1 \Rightarrow Y) \;=\; \frac{\Var(W_1' - W_1)}{2\,\Var(Y(W))} \;=\; \frac{1}{2}.
  \]
  By ancestral consistency (\Cref{thm:ancestral-consistency}), this discrepancy remains true if we marginalize out $W_2$ (so the graph is $W_1 \rightarrow Y$ and the potential outcomes of $Y$ are given by $Y(w_1) = w_1 + w_1 E_2$ in Model 1 and $Y(w_1) = w_1 + E_2$ in Model 2).
\end{example}

Heuristically, the partial identification problem occurs because
counterfactual explainability depends on the joint probability
distribution of potential outcomes schedule, but we can, at the very
best, obtain the marginal distribution of ``single-world'' potential
outcomes using empirical data. In the example above, we can estimate
the marginal distributions of \(W_2(w_1 = 1)\) and \(W_2(w_1 = -1)\)
empirically, but no data can allow us to identify the copula of these
two variables under different interventions. This challenging partial
identification problem does not occur in global sensitivity analysis
because it assumes that the outcome variable is a deterministic
function of the explanatory variables.

We will not address the partial identification problem for
counterfactual explainability in this paper. Instead, we impose a
comonotonicity assumption on the
potential outcomes under which counterfactual explainability is point
identifiable. The comonotonicity assumption is sometimes called rank
preservation in the literature
\citep{robins1989probability,hernanCausalInferenceWhat2023} and is
commonly used in counterfactual problems for this purpose \citep[see
e.g.][]{heckman1997making,lu2023evaluating}. Of course, this does not
justify the assumption in practical applications; see
\Cref{sec:discussion} for some discussion.

\begin{definition}[Comonotonicity of all basic potential
  outcomes]\label{defi:comonotonicity}
  Consider a causal Markov distribution with respect to a DAG $G$ with
  vertex set $V = (W_1,\dots,W_K,Y)$ and a fixed anchor $v$ in the
  support of $V$. We say the basic potential outcomes are
  \emph{comonotone} if there exist
  non-decreasing functions $h_{k,v'}$ for $k \in [K+1]$ and $v'$ in
  the support of $V$ such that $V_k(v'_{\pa(k)}) =
  h_{k,v'}(V_k(v_{\pa(k)}))$ for all $k$ and $v'$ with probability $1$.
\end{definition}

In making this assumption, it is implicitly assumed that
$W_1,\dots,W_K$ are real-valued. Comonotonicity essentially requires
that any two basic potential
outcomes of the same variable are perfectly positively correlated.
In Example \ref{exam:identification with dependent inputs}, Model 2 is
comonotone because $Y(w_1 = 1) = Y(w_1 = -1) + 2$ and $x \mapsto x +
2$ is an increasing function, and Model 1 is not comonotone because $Y(w_1
= 1) = -Y(w_1 = -1)$ and $x \mapsto -x$ is decreasing. As another
example, an NPSEM-IE model (Example \ref{exam:NPSEM}) is comonotone if
the noise is additive, that is, if $V_k(v_{\text{pa}}(k)) = f_k(v_{\text{pa}(k)}, E_k) =
g_k(v_{\text{pa}(k)}) + E_k$ for all $k$, because $x \mapsto x +
g_k(v'_{\pa(k)}) - g_k(v_{\pa(k)})$ is increasing.

Comonotonicity allows us to determine the joint distribution of the
basic potential outcomes using their marginal distributions, which can
be estimated from empirical data. Specifically, let $Q_k(e \mid
v_{\pa(k)})$ ($e \in [0,1]$) denote the conditional quantile function of $V_k$
given $V_{\pa(k)} = v_{\pa(k)}$. The observable data distribution can
then be described by a sequence of conditional quantile
transformations:
\begin{equation} \label{eq:conditional-quantile}
  V_k = Q_k(E_k \mid V_{\pa(k)}),~E_k \overset{i.i.d.}{\sim}
  \text{Unif}[0,1],\quad k \in [K+1].
\end{equation}
Under comonotonicity, the joint distribution of the potential outcomes
schedule is the same as that obtained by treating the above equations
as structural, that is by assuming
\begin{equation}
  \label{eq:comonotone-coupling}
  V_k(v_{\pa(k)}) = Q_k(E_k \mid v_{\pa(k)}),\quad k \in [K+1].
\end{equation}
Indeed, it is easy to verify that this defines a comonotone
distribution using
$F_k(\,\cdot\,\mid V_{\pa(k)}=w'_{\pa(k)}) \circ Q_k(\,\cdot\,\mid
V_{\pa(k)}=v_{\pa(k)})$ as the the map $h_{k,w'}$ in
Definition \ref{defi:comonotonicity}, where $F_k$ denotes the cumulative
distribution function of $V_k$ given $V_{\pa(k)} = v_{\pa(k)}$.
Following \Cref{exam:NPSEM}, we can treat $E = (E_1,\dots,E_K)$ in
\eqref{eq:conditional-quantile} as causally independent noise
variables and $Y$ as a function of $E$ in an NPSEM-IE. This leads to
the next identifiability result.


\begin{theorem}[Identifiability under
  comonotonicity]\label{thm:identification.comonotone}
  Suppose the distribution of the potential outcomes schedule of
  $V = (W_1,\dots,W_K,Y)$ is causal Markov with respect to a DAG $G$ and the
  basic potential outcomes are comonotone. Then the counterfactual
  explainability $\xi_{\gG}(\cdot \Rightarrow Y)$ can be uniquely identified
  by the joint distribution of $V$.
\end{theorem}


\subsection{Estimation}
\label{sec:estimation-1}

We next discuss estimation of counterfactual explainability $\xi_G$ under
comonotonicity. Given the conditional quantile functions $Q_k(\cdot
\mid v_{\pa(k)}), k \in [K+1]$ which define the comonotone NPSEM-IE
via \eqref{eq:comonotone-coupling}, we can compute the total
counterfactual explainability $\xi_G(\vee_{k \in \calS} V_k
\Rightarrow Y)$ for any set $\calS \subseteq [K+1]$ as defined in
\eqref{eq:xi-g}. More specifically, we can rewrite \eqref{eq:xi-g} as
\[
  \xi_{G,\P}(\vee_{k \in \calS} V_k \Rightarrow Y) =
  \frac{\var\left(f(V^*) - f(V^{*\prime}_{\calS},
      V^*_{-\calS})\right)}{\var\left(f(V^*) -
      f(V^{*\prime})\right)}.
\]
For the numerator, we sample $f(V^{*})$ by sampling independent
uniform variables $E_1,\dots,E_{K+1}$ and evaluating
\eqref{eq:conditional-quantile} sequentially. We can sample
$f(V^{*\prime}_{\calS},
V^*_{-\calS})$ analogously with $E_k$ in \eqref{eq:comonotone-coupling}
in with an i.i.d.\ copy $E_k'$ for all $k \in \calS$. We then estimate
the numerator by averaging $(f(V^{*}) - f(V^{*\prime}_{\calS},
V^*_{-\calS}))^2$ over the sample. We estimate
the denominator in the display equation above analogously using Monte
Carlo (without ``freezing'' $E_{-\calS}$). Counterfactual
explainability of other clauses in $\calE(V)$ can then be obtained by
the inclusion-exclusion principle.

Thus, the key step is to estimate the conditional quantile functions
from data. To this end, we will consider three methods:

\begin{enumerate}
\item Suppose the structural equations have additive noise,
  so $V_k(v_{\pa(k)}) = g_k(v_{\pa(k)}) + E_k$. We can apply
  nonparametric regression methods to obtain an estimator
  $\hat{g}_k(v_{\pa(k)})$ for $g_k(v_{\pa(k)}) =
  \E(V_k \mid V_{\pa(k)} = v_{\pa(k)})$ and use the empirical
  distribution of the residuals to estimate the law of $E_k$; that
  is, when generating the counterfactual we randomly sample a
  residual and add it to the estimated conditional mean.
\item Often, the distribution of the residual $R_k =
  V_k(v_{\pa(k)}) - \E(V_k \mid V_{\pa(k)} = v_{\pa(k)})$ may
  depend on $v_{\pa(k)}$. In this case, we assume $R_k$ is a
  mean-zero Gaussian random variable with variance
  $\sigma_k^2(v_{\pa(k)})$ and estimate the variance function by
  regressing the squared empirical residuals $(V_k -
  \hat{g}_k(V_{\pa(k)}))^2$ on $V_{\pa(k)}$. The counterfactuals
  can then be sampled sequentially by generating new residuals
  from the estimated Gaussian distribution.
\item Most generally, we can apply general quantile
  regression methods to estimate the conditional quantiles
  $Q_k(\tau \mid v_{\pa(k)}), k \in [K+1]$ for a grid of levels
  $\tau \in [0,1]$ (e.g.\ with XGBoost with quantile loss
  functions) and interpolate between the grid points to estimate
  the conditional quantile functions. The counterfactuals can then
  be sampled sequentially using \eqref{eq:comonotone-coupling}.
\end{enumerate}


\section{Real data example: Explaining income inequality}\label{sec:simulation}

\subsection{Background: datasets, causal DAGs, and
  statistical methods}\label{sec:data}

We demonstrate the concept of counterfactual explainability and the
estimation methods in \Cref{sec:estimation-1} using the UCI
\texttt{Adult} dataset of U.S. working adults in 1994 and the 2018
\texttt{ACSIncome} dataset
\citep{ding2021retiring}.\footnote{We used the 1994 \texttt{Adult}
  dataset reconstructed by \citet{ding2021retiring}. The
  \texttt{ACSIncome} data spans from 2014 to 2018, and we only
  consider 2018. As the data comprise over $1.6$ million records, we
  randomly subsampled $50,000$ observations to make the sample size
  comparable to that of the UCI \texttt{Adult} dataset.}.
In this
example, the variable to be
explained is the logarithm of the annual income (US dollars, minimal
value \$100). (The log transform is used because the raw income is
heavily right-skewed.) Explanatory factors include sex (Female;
Male), race (White; Black/African American; Asian or Native Hawaiian and
Other Pacific Islander; American Indian or Alaska Native; Other), and
education (an integer from $1$ to $24$ that approximates
the completed years of schooling). Our analysis below will be restricted to
$36,566$ adults in the dataset who were at least $25$ years old and
born in the United States.  
To examine how explanations of income inequality change over time, the
sample is partitioned into $8$ age bins (see
\Cref{tab:age.group}).

\begin{table}[btp]
  \centering
  \caption{Number of
    individuals by age group in the UCI \texttt{Adult} dataset.}\label{tab:age.group}
  \begin{tabular}{l*{8}{c}}
    \toprule
    Age group & [25,30) & [30,35) & [35,40) & [40,45) & [45,50) & [50,55) & [55,60) & Over $60$ \\
    \midrule
    Count & 5404 & 5826 & 5860 & 5223 & 4507 & 3449 & 2563 & 3734 \\
    \bottomrule
  \end{tabular}
\end{table}


A key feature of the counterfactual explainability $\xi_G$ is the
consideration of causal dependencies between explanatory and
outcome variables through a pre-specified DAG $G$. We will use the DAG
in \Cref{fig:real.data}(a) to estimate counterfactual
explainability. In
this graph, sex and race are root vertices with no parents as they
are fixed at conception, education may be causally influenced by
sex and race, and income may be influenced by sex, race, and
education. Thus, sex can influence income through two pathways,
including an indirect pathway through educational attainment. We also
consider two ``marginalizations'' of this
graph in \Cref{fig:real.data}(b) and (c) to demonstrate the ancestral
consistency property of counterfactual explainability.
We will assume comonotonicity of the basic potential outcomes
(Definition \ref{defi:comonotonicity}), so the counterfactual
explainabilities are point identified by Theorem
\ref{thm:identification.comonotone}.

\begin{figure}[t] \centering
  \begin{minipage}[b]{0.3\textwidth} \centering
    \begin{tikzcd}
      \text{Race} \arrow[d, blue]\arrow[dr, blue] & \\
      \text{Education} \arrow[r, blue]&\text{Income} \\
      \text{Sex} \arrow[u, blue] \arrow[ur, blue]  &
    \end{tikzcd}
    \caption*{(a) }
  \end{minipage}
  \begin{minipage}[b]{0.3\textwidth} \centering
    \begin{tikzcd}
      \text{Race} \arrow[dr, blue] & \\
      & \text{Income} \\
      \text{Sex} \arrow[ur, blue] &
    \end{tikzcd}
    \caption*{(b) }
  \end{minipage}
  \begin{minipage}[b]{0.3\textwidth} \centering
    \begin{tikzcd}
      \\
      & \text{Income} \\
      \text{Sex} \arrow[ur, blue] &
    \end{tikzcd}
    \caption*{(c) }
  \end{minipage}

  \caption{DAGs for explaining annual income.}
  \label{fig:real.data}
\end{figure}

For our main analysis, we will use nonparametric estimates of the
conditional quantile functions (third method in
\Cref{sec:estimation-1}) as education and log income appear to have
skewed/multi-modal distributions with heterosckedasticity (see
\Cref{sec:distribution}). We next describe the
conditional quantile-based estimation procedure in more detail. For the two
root nodes (race and sex), we estimated cell probabilities by
the empirical frequencies and sampled these variables accordingly. For
the conditional quantile of education given race and sex, we 
used XGBoost \citep{chen2016xgboost} with the quantile loss on a grid
of levels $\{0.01, 0.03, \ldots,0.97, 0.99\}$ with 
isotonic rearrangement to enforce monotonicity of the quantile functions.
We then approximated the entire conditional quantile function by
piecewise–linear interpolation and clamped outside the grid (the
functiona is flat at the boundary) to avoid
unstable tail extrapolation. Conditional quantiles of log income given
race, sex, and education are estimated analogously. We use 1000
Monte Carlo repeats to estimate all the total counterfactual
explainabilities and then use the inclusion-exclusion principle to
determine the explainability of all other clauses.


\subsection{Results}\label{sec:results}

\begin{figure} \centering
  \begin{tabular}{c c c}
    Age & Income in 1994 & Income in 2018 \\
    $[25,30)$ &
                \begin{minipage}{.27\textwidth}
                  \includegraphics[clip, trim = 1cm 2cm 0.8cm 2.5cm, width = \textwidth]{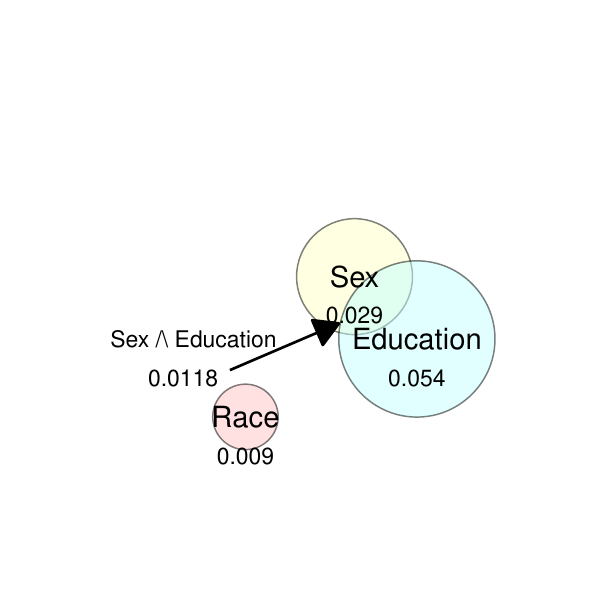}
                \end{minipage}
                         &
                           \begin{minipage}{.27\textwidth}
                             \includegraphics[clip, trim = 1cm 2cm 0.8cm 2.5cm, width = \textwidth]{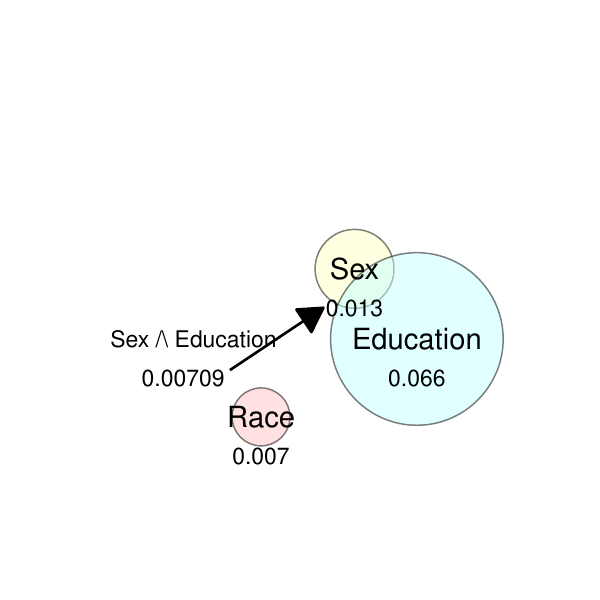}
                           \end{minipage} \\
    $[30,35)$ &
                \begin{minipage}{.27\textwidth}
                  \includegraphics[clip, trim = 1cm 2cm 0.8cm 2.5cm, width = \textwidth]{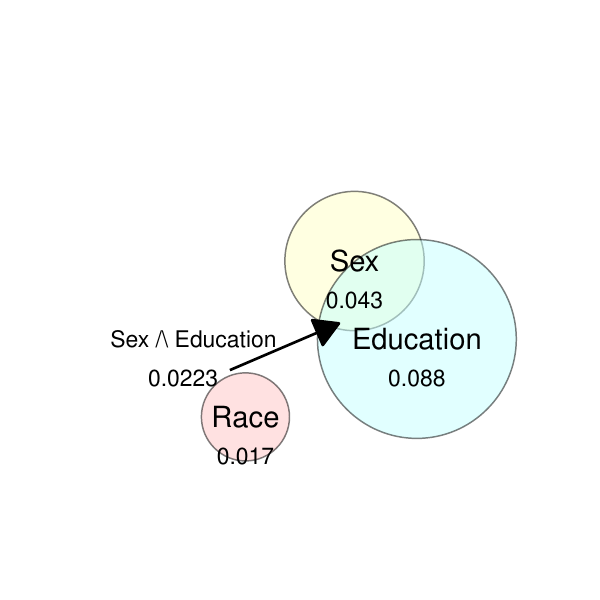}
                \end{minipage}
                         &
                           \begin{minipage}{.27\textwidth}
                             \includegraphics[clip, trim = 1cm 2cm 0.8cm 2.5cm, width = \textwidth]{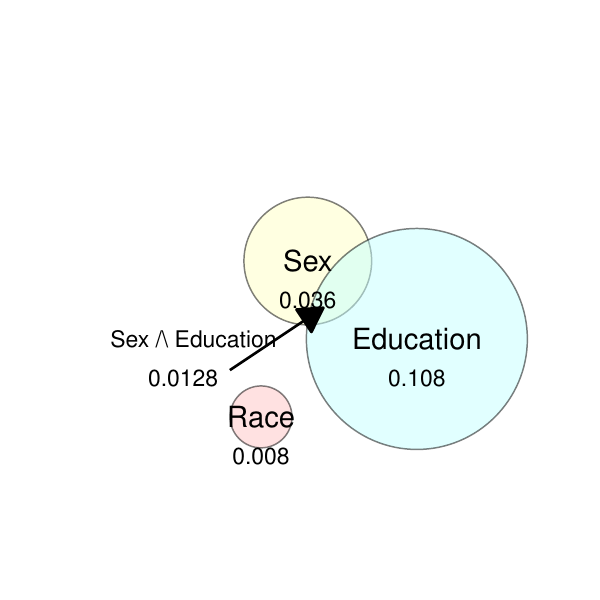}
                           \end{minipage} \\
    $[35,40)$ &
                \begin{minipage}{.27\textwidth}
                  \includegraphics[clip, trim = 1cm 2cm 0.8cm 2.5cm, width = \textwidth]{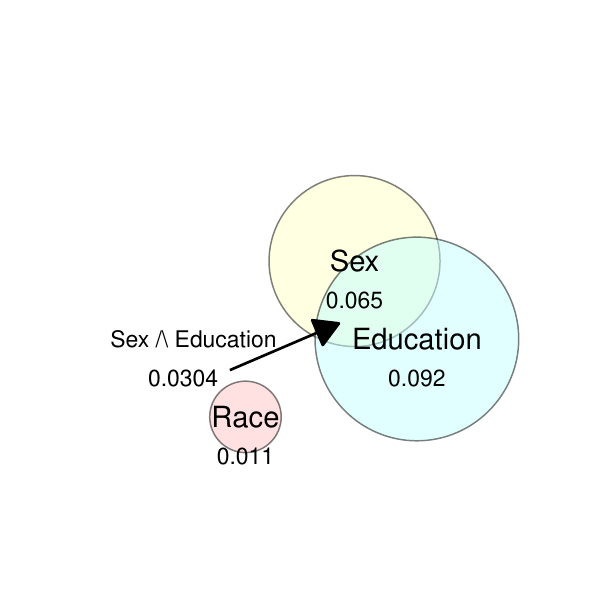}
                \end{minipage}
                         &
                           \begin{minipage}{.27\textwidth}
                             \includegraphics[clip, trim = 1cm 2cm 0.8cm 2.5cm, width = \textwidth]{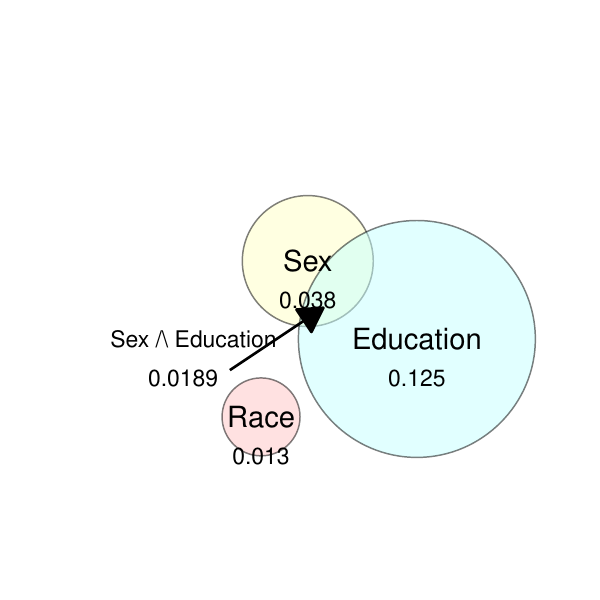}
                           \end{minipage} \\
    $[40,45)$ &
                \begin{minipage}{.27\textwidth}
                  \includegraphics[clip, trim = 1cm 2cm 0.8cm 2.5cm, width = \textwidth]{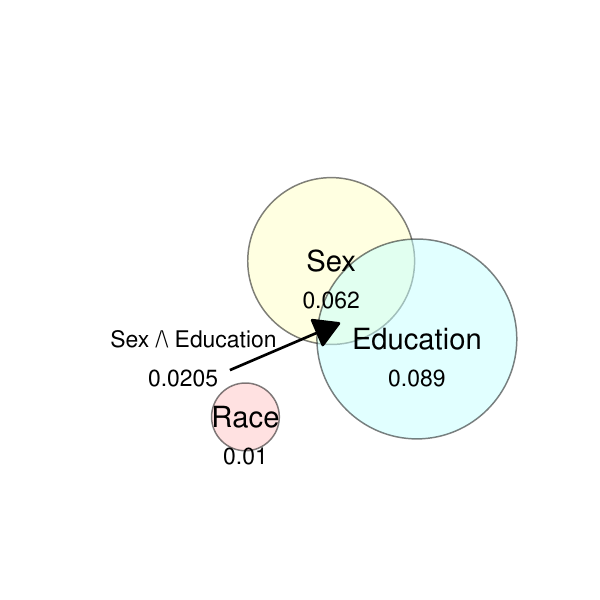}
                \end{minipage}
                         &
                           \begin{minipage}{.27\textwidth}
                             \includegraphics[clip, trim = 1cm 2cm 0.8cm 2.5cm, width = \textwidth]{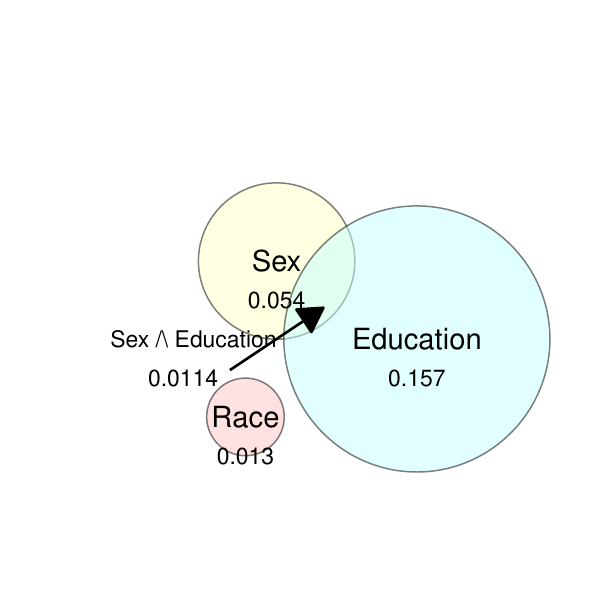}
                           \end{minipage} \\
    $[45,50)$ &
                \begin{minipage}{.27\textwidth}
                  \includegraphics[clip, trim = 1cm 2cm 0.8cm 2.5cm, width = \textwidth]{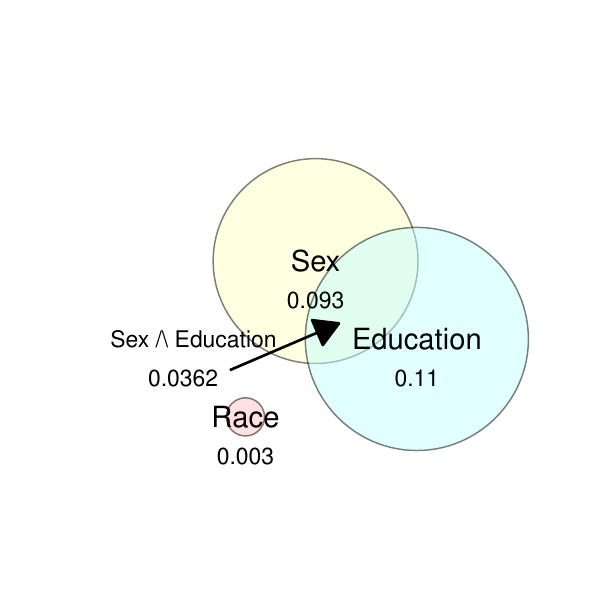}
                \end{minipage}
                         &
                           \begin{minipage}{.27\textwidth}
                             \includegraphics[clip, trim = 1cm 2cm 0.8cm 2.5cm, width = \textwidth]{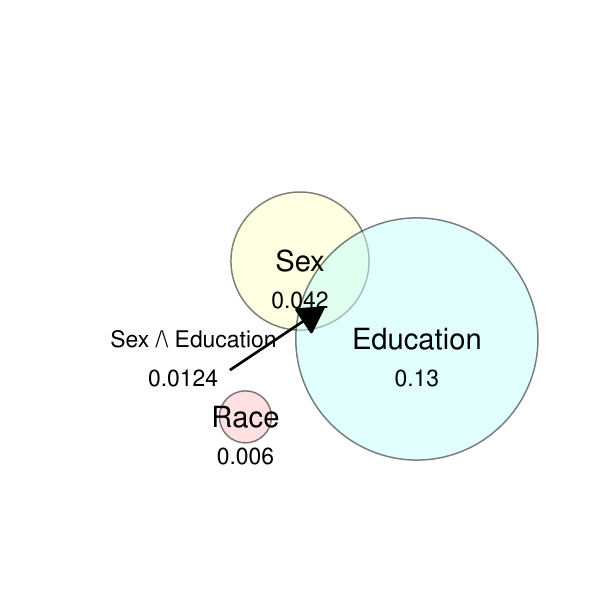}
                           \end{minipage} \\
    $[50,55)$ &
                \begin{minipage}{.27\textwidth}
                  \includegraphics[clip, trim = 1cm 2cm 0.8cm 2.5cm, width = \textwidth]{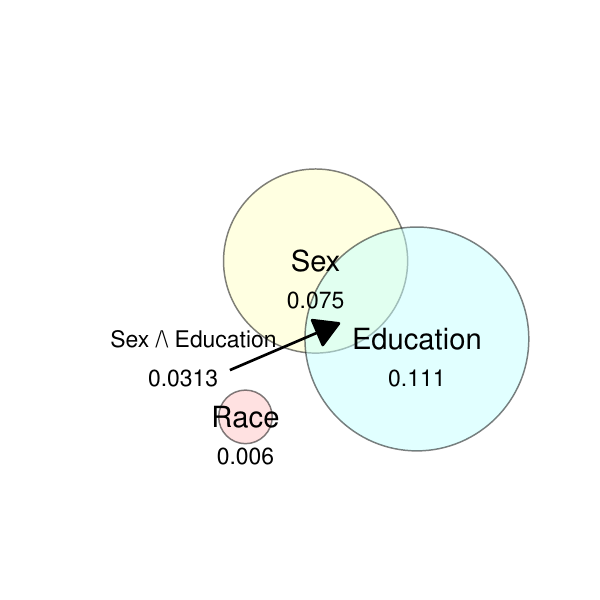}
                \end{minipage}
                         &
                           \begin{minipage}{.27\textwidth}
                             \includegraphics[clip, trim = 1cm 2cm 0.8cm 2.5cm, width = \textwidth]{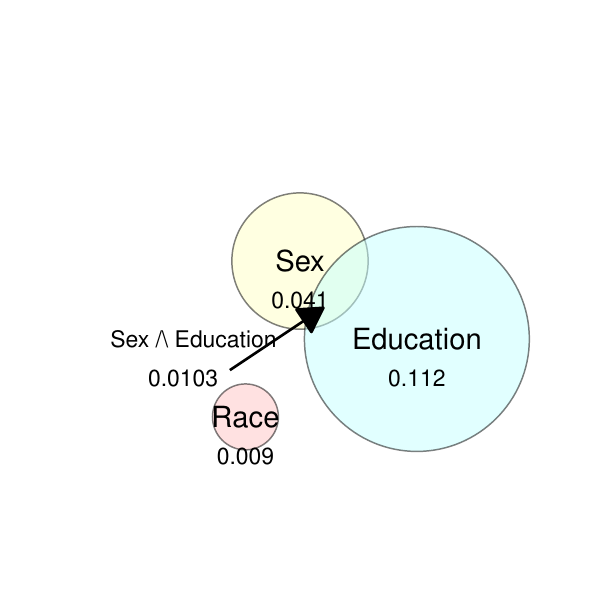}
                           \end{minipage} \\
    $[55,60)$ &
                \begin{minipage}{.27\textwidth}
                  \includegraphics[clip, trim = 1cm 2cm 0.8cm 2.5cm, width = \textwidth]{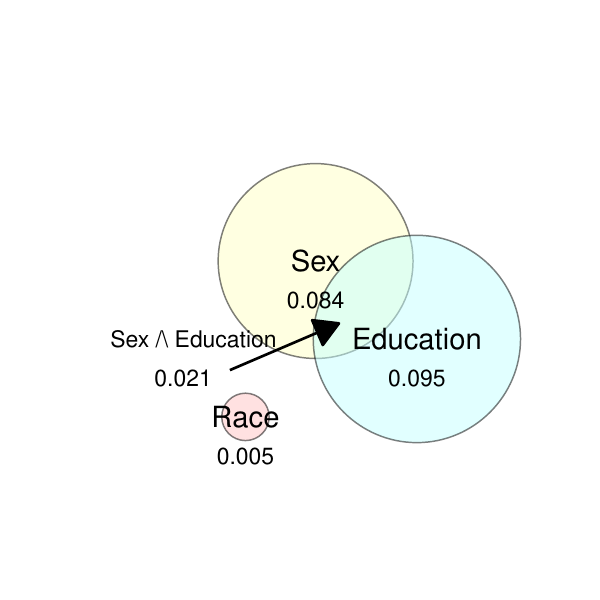}
                \end{minipage}
                         &
                           \begin{minipage}{.27\textwidth}
                             \includegraphics[clip, trim = 1cm 2cm 0.8cm 2.5cm, width = \textwidth]{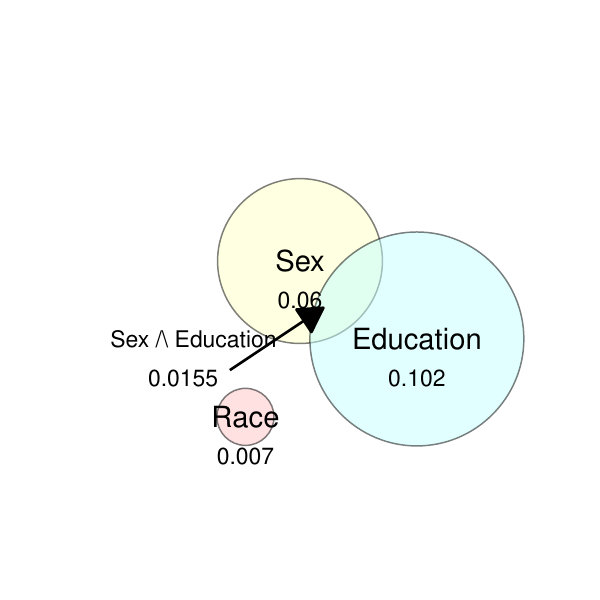}
                           \end{minipage} \\
    Over $60$ &
                \begin{minipage}{.27\textwidth}
                  \includegraphics[clip, trim = 1cm 2cm 0.8cm 2.5cm, width = \textwidth]{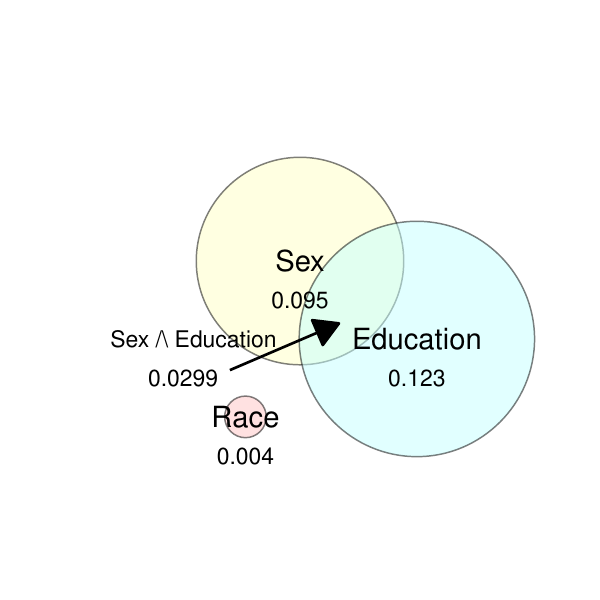}
                \end{minipage}
                         &
                           \begin{minipage}{.27\textwidth}
                             \includegraphics[clip, trim = 1cm 2cm 0.8cm 2.5cm, width = \textwidth]{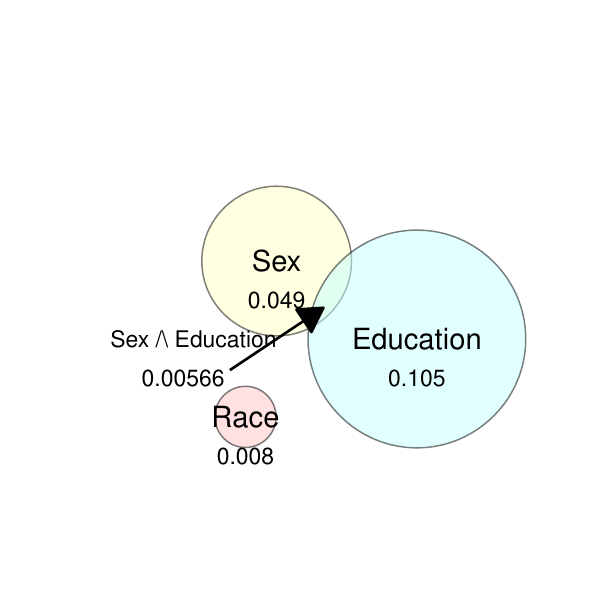}
                           \end{minipage}
  \end{tabular}
  \caption{Venn diagrams for the estimated explainabilities from the
    US Income dataset across age groups.}
  \label{fig:real.data.across.age.bin}
\end{figure}

The Venn diagrams in \Cref{fig:real.data.across.age.bin} visualize the
estimated counterfactual explainability for different age groups in
1994 and 2018.
In most cases, the total explainability of
race, sex, and education is between $10\%$ and $15\%$. Among single
factors, education has the largest total explainability
($5.4\%$ to $15.7\%$ across age groups), followed by sex ($1.3\%$
to $9.3\%$ across age groups), and race explains little to
none. The Sex $\times$ Education
interaction has non-zero explainability across all age groups, taking
up about $1/3$ to $1/2$ of sex's total explainability. This indicates
that the economic return of education appears to differ between men
and women.

An interesting observation is that the estimated counterfactual
explainability appear to vary considerably across age groups in in
year 1994 and 2018, reflecting generational differences in income
inequality. In particular, the total
explainability of sex is much smaller in the younger population,
especially those between $25$ and $30$, than older age groups. In
addition, the total explainability of sex is smaller in 2018 than in
1994. These patterns suggest that the income inequality by
gender is narrowing over time, possibly due to the social changes that
expanded women's access to higher education and career
opportunities. The total explainability of educational attainment also
demonstrates some interesting patterns in
\Cref{fig:real.data.across.age.bin}. In particular, in 1994 education
explains the most in the oldest age group while in 2018 it explains
the most in the mid-career 40--45 group.



\begin{figure}[t]
  \centering
  \begin{minipage}{\textwidth}
    \centering
    \includegraphics[clip, trim = 0cm 0cm 0cm 0cm, width = 0.3\textwidth]{plot/incomeResultConditionalQuantileOver60.pdf}
  \quad
    \includegraphics[clip, trim = 0cm 0cm 0cm 0cm, width = 0.3\textwidth]{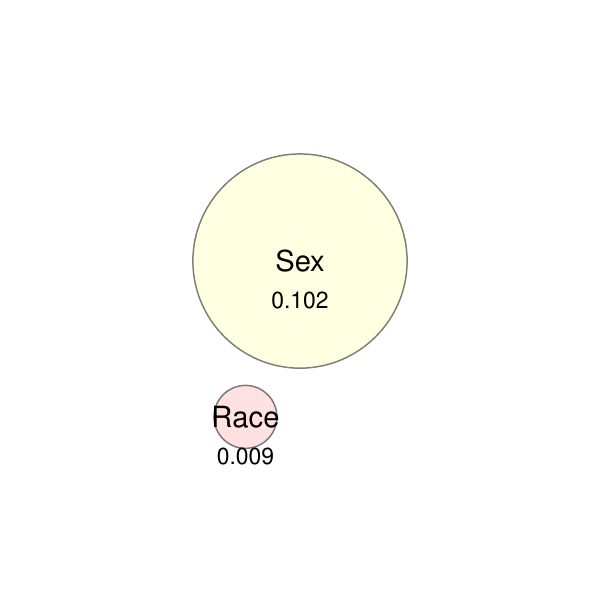}
  \quad
    \includegraphics[clip, trim = 0cm 0cm 0cm 0cm, width =
    0.3\textwidth]{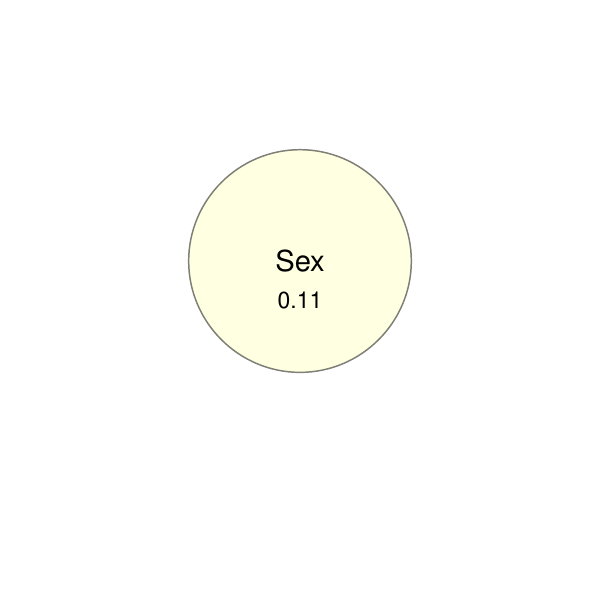}
    \caption*{(a) Estimated counterfactual explainabilities in 1994
      for the oldest age group (over 60) using the three DAGs in
      \Cref{fig:real.data}.}
  \end{minipage}
  \begin{minipage}{1\textwidth}
                \centering
                \includegraphics[clip, trim = 0cm 0cm 0cm 0cm, width =
                1\textwidth]{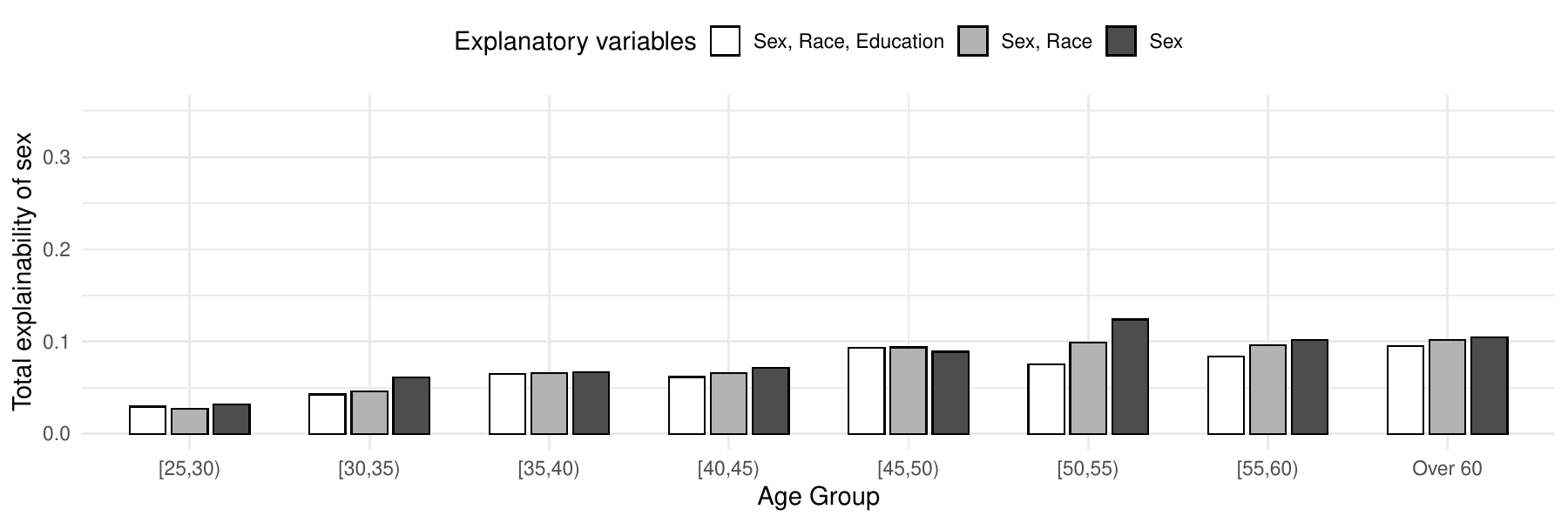}
  \caption*{(b) Estimated total explainability of
      sex to log annual income in 1994 in different age groups using
      the three DAGs in \Cref{fig:real.data}.}
  \end{minipage}
  \caption{Demonstration of the ancestral consistency property of
    counterfactual explainability.} \label{fig:partial}
\end{figure}

As shown in \Cref{thm:ancestral-consistency}, counterfactual
explainability has certain consistency properties with respect to the
ancestral margins of the causal DAG. To illustrate this, we consider
counterfactual explainabilities for the three graphs in
\Cref{fig:real.data} (the same conditional quantile-based approach is
used for the smaller graphs in panels (b) and (c)). Because sex is
a root node in all three DAGs, by \Cref{thm:ancestral-consistency} its total explainabilities should be the
same. Such consistency property is demonstrated in
\Cref{fig:partial}. Although the numerical values are not identical
because the estimated conditional quantile functions may not be
consistent, the explainabilities obtained from different DAGs are
broadly consistent.


We provide a number of additional results in the Appendix:
\begin{enumerate}
\item Exploratory data analysis. In \Cref{sec:distribution}, we assess
  the suitability of the three estimation procedures outlined in
  \Cref{sec:estimation-1} by investigating the conditional
  histograms of education and log income.
\item Estimates by other methods in \Cref{sec:estimation-1}. We add
  the results of the other two
  estimation procedures in \Cref{sec:other.estimation},
  which are considerably different with the approach taken above. This
  highlights the importance of using appropriate methods to estimate
  the conditional quantile functions.
\item Comparison with Shapley's value. We further compare
the total counterfactual explainability of sex with Shapley's value
of sex in \Cref{sec:comp-with-shapl}. By definition, Shapley's value
depends on what other explanatory variables are included, and we
indeed observe more instability from Shapley's value in this real data
example.
\item Partial identification bounds. As discussed in
  \Cref{sec:estimation}, without further assumptions  counterfactual
  explainability can only partially identified using empirical
  data. We demonstrate this in \Cref{sec:partial.ID} using the bounds
  for total explainability obtained in
  \citet{leiHeritabilityCounterfactualPerspective2025}.
\end{enumerate}

\section{Discussion}\label{sec:discussion}



Our approach in this article can be seen as an extension of the
classical approach by Sobol' and others for global sensitivity analysis
based on functional ANOVA in several directions:
\begin{enumerate}
\item we introduce the concept of explanation algebra to unify several
  notions of variable importance and visualize the results; 
\item
  we provide a simple recipe for defining and estimating the
  explainability of dependent explanatory variables using the causal
  DAG theory;
\item counterfactual explainability can be applied to understand
  generative models in very much the same way as how global
  sensitivity analysis has been applied to understand deterministic
  prediction models.
\end{enumerate}
Exanding the last point above, in making the comonotonicity assumption
(Definition \ref{defi:comonotonicity}) it is implicitly assumed that
all explanatory variables $W_1,\dots,W_K$ are real-valued. But this is
not needed in defining counterfactual explainability. In fact,
counterfactual explainability can be computed using the approach in
\Cref{sec:estimation-1} even for non-Euclidean
$W_1,\dots,W_K$ as long as their counterfactuals can be sampled from a
generative model.

It is possible to further extend our approach to allow exogenous
correlation (unmeasured confounding) between the explanatory variables
by using acyclic directed mixed graphs,
in which the exogenous correlations are represented by bidirected
edges
\citep{pearlTheoryInferredCausation1991,richardson03_markov_proper_acycl_direc_mixed_graph,zhaoStatisticalCausalModels2025}. In
this case, one
can treat the set of variables connected by bidirected edges (this is
often called a \emph{district} in the graphical models literature) as
one ``big explanatory factor'' in forming the explanation
algebra, as it is generally not possible to distinguish the
explanation due to interaction and exogenous correlation within a
district.

Although the counterfactual extension is conceptually clean, a
practical challenge is that counterfactual explainability is generally
only partially identified because it depends on the distribution of
cross-world potential outcomes. We have sidestepped this issue by
using the comonotone coupling (\Cref{defi:comonotonicity}), but the
comonotonicity assumption may
not hold in applications. When a subset of explanatory variables
$W_{\calS}$ has no ``gaps'' in the causal DAG (there exists no chain
$W_{k_1} \rightarrow W_{k_2} \rightarrow W_{k_3}$ such that $k_1,k_3
\in \calS$ but $k_2 \not \in \calS$), the comonotone coupling provides
tight lower bound for total explainability. This result follows from
the classical Frech\'{e}t-Hoeffding copula bound; see
\citet{leiHeritabilityCounterfactualPerspective2025}. But in general,
there are no closed-form tight upper bounds for total explainability
or closed-form bounds for the counterfactual explainability of other
clauses (such as interaction explainability). Solving such partial
identification problems may require developing computational tools using
multi-marginal optimal transport \citep{gao2024bridge}.

\acks{QZ was partly funded by EPSRC grant EP/V049968/1. The authors thank Hongyuan Cao, Haochen Lei, Jieru (Hera) Shi, and Art Owen for helpful discussion.}

\bibliography{ref}


\appendix
\section{Proofs}
\label{app:proofs}

\subsection{Proof of \Cref{thm:xi-equivalent}}

\begin{proof}
  It is easy to see that $(\wedge_{k \in \calS} W_k) \wedge (\wedge_{k
    \notin \calS} \neg W_k)$ is mapped to the singleton
  $\{1_{\calS}\}$ in the set algebra, where $1_{\calS}$ is the
  indicator vector for $\calS$ ($w_k = 1$ if $k \in \calS$ and $w_k =
  0$ if $k \not \in \calS$). It follows from the ANOVA decomposition
  $\var(f(W)) = \sum_{\calS \subseteq [K]} \sigma_{\calS}^2$ that
  \eqref{eq:xi-anova} and finite additivity defines a probability
  measure $\xi_1$ on $\mathcal{E}(W)$.

  To show that $\xi_1$ must be the same as the other probability
  measures, recall Dynkin's $\pi$-$\lambda$ theorem that says two
  probability measures on the same $\sigma$-algebera must be the same
  if they agree on a $\pi$-system \citep[Theorem
  3.3]{billingsleyProbabilityMeasure2012}; a collection of events is a
  $\pi$-system if it is non-empty and closed under intersection. It is
  easy to see that the collections of explanation clauses in
  \eqref{eq:xi-sobol-lower} and \eqref{eq:xi-superset} are
  $\pi$-systems. And although the collection in
  \eqref{eq:xi-sobol-upper} is not a $\pi$-system, the collection of
  the negation of all the clauses in \eqref{eq:xi-sobol-upper} is the
  same as that in \eqref{eq:xi-superset} and thus is a
  $\pi$-system. Moreover, it is easy to verify using the definitions of
  $\underline{\tau}^2_{\calS}$, $\overline{\tau}^2_{\calS}$, and
  $\overline{\sigma}^2_{\calS}$ that $\xi_1$ agrees with the
  other measures on the corresponding clauses in
  \eqref{eq:xi-sobol-lower}, \eqref{eq:xi-sobol-upper}, and
  \eqref{eq:xi-superset}. For example, we have
  \[
    \vee_{k \in \calS} W_k = \vee_{\calS' \cap \calS \neq \emptyset} \{
    (\wedge_{k \in \calS'} W_k) \wedge (\wedge_{k \notin \calS'} \neg
    W_k)\},
  \]
  whereas, by definition, $\overline{\tau}^2_{\calS} = \sum_{\calS'
    \cap \calS \neq \emptyset} \sigma^2_{\calS'}$. Thus the conclusion
  is a simple consequence of the $\pi$-$\lambda$ theorem.
\end{proof}

\subsection{Proof of \Cref{thm:uniqueness}}

\begin{proof}
  Denote $\zeta_Y(Y \Rightarrow Y)$ as $\zeta(Y)$, which is a function
  of the distribution of $Y$. Totality and Continuity in
  Definition \ref{defi:axiom} imply that $\zeta(Y)$ is non-negative and
  continuous. By further using Linearity, Symmetry, and Additivity I,
  we have, for $W = (W_1,W_2) \in L_{r,\indep}^2$ and $Y = W_1 + W_2$,
  \begin{align*}
    \zeta(Y) &= \zeta_{W}(W_1 \vee W_2 \Rightarrow W_1 + W_2) \\
             &= \zeta_W(W_1 \Rightarrow W_1) + \zeta_W(W_2 \Rightarrow W_2) \\
             &= \zeta_{W_1}(W_1 \Rightarrow W_1) + \zeta_{W_2}(W_2 \Rightarrow W_2) \\
             &= \zeta(W_1) + \zeta(W_2).
  \end{align*}
    By further using Totality, Linearity, Symmetry, and Additivity I,
    we have, for $W = (W_1,W_2) \in L_{r,\indep}^2$ and $Y = W_1 + W_2$,
    \begin{align*}
      \zeta(Y) &= \zeta_{W}(W_1 \vee W_2 \Rightarrow W_1 + W_2) \\
               &= \zeta_W(W_1 \vee W_2 \Rightarrow W_1) + \zeta_W(W_1 \vee W_2 \Rightarrow W_2) \\
               &= \int \zeta_{W_{-2}}(W_1 \Rightarrow W_1) d \P(w_2)+ \int \zeta_{W_{-1}}(W_2 \Rightarrow W_2) d \P(w_1) \\
               &=  \int \zeta(W_1 \Rightarrow W_1) d \P(w_2)+ \int \zeta(W_2 \Rightarrow W_2)d \P(w_1) \\
               &= \zeta(W_1) + \zeta(W_2).
    \end{align*}
  This shows that $\zeta(Y)$ is additive with respect to convolution
  of the distributions. By \citet[1.12 and
  1.14]{mattner1999cumulants}, there exists $c > 0$ such that
  \begin{equation}
    \label{eq:zeta-total}
    \zeta(Y) = c\Var(Y).
  \end{equation}
  Now by using Linearity, Totality, and \eqref{eq:zeta-total}, we have
  \begin{align*}
    \zeta_W(\vee_{k \in [K] \setminus \{j\}} W_k \Rightarrow Y) &= \int
                                                                  \zeta_{W_{-j}} \left( \vee_{k \in [K] \setminus \{j\}} W_k \Rightarrow
                                                                  Y(w_j)  \right) \diff \P(w_j) \\
                                                                &= \int c \Var(Y(w_j)) \diff \P(w_j) \\
                                                                &= c \E\{\Var(Y \mid W_j)\}.
  \end{align*}
  By applying this argument recursively, we obtain the first claim in
  the Theorem. Finally, if $\zeta$ additionally
  satisfies Additivity II, it is a measure on $\mathcal{E}(W)$. By
  normalizing $\zeta_W(\cdot \Rightarrow Y)$ using the total measure,
  we obtain the probability measure $\xi_3$ in
  \Cref{thm:xi-equivalent}.
\end{proof}

\subsection{Proof of \Cref{thm:ancestral-consistency}}
\label{sec:proof-crefthm:-cons}

\begin{proof}
  The first conclusion follows from \eqref{eq:xi-ancestral} and the
  fact that $Y(w_{\calS}) = Y(w_{\calS'})$ because all directed paths
  from $W_{\calS'} \setminus W_{\calS}$ to $Y$ must go through
  $W_{\calS}$. The second conclusion follows from an argument similar to that in the proof of \Cref{thm:xi-equivalent} with the definition of $\xi_G$ and the fact that basic potential
  outcomes $W_j^{*}$, $j \in \calS$ are the same in the causal Markov
  models with respect to $G$ and $G'$.
\end{proof}

\subsection{Proof of \Cref{thm:identification.comonotone}}

\begin{proof}
  Equation \eqref{eq:comonotone-coupling} defines a NPSEM-IE and the
  quantile functions $Q_k(\cdot \mid v_{\pa(k)})$ can be identified
  from the joint distribution of $V$, so the
  total explainability $\xi_G \left( \vee_{k \in
      \calS} W_{k} \right)$ for any $\calS \subseteq [K+1]$ is
  identified by \eqref{defi:total.dependent.NPSEM}. The entire
  explainability measure on the algebra $\mathcal{E}(W)$ can
  subsequently be identified using probability calculus.
\end{proof}

\section{Defining explainability measure using
  $\xi_3$} \label{sec:inclusion-exclusion}

We have established the following result in the main text.

\begin{theorem} \label{thm:total-explainability-extension}
  The definition of total explainability in equation
  \eqref{eq:xi-sobol-upper} extends to a unique probability measure
  $\xi_3$ on $\mathcal{E}(W)$
\end{theorem}

In particular, functional ANOVA establishes the existence and
\Cref{thm:xi-equivalent} establishes uniqueness of $\xi_3$.

In this Appendix, we give an alternative proof of this fact using the
anchored decomposition of a function. Combined with the axiomatization
of total explainability in \Cref{sec:axiom-total-expl}, this provides a
justification of the explainability measure $\xi$ without using the
Hoeffding/functional ANOVA decomposition. To this end, we first
introduce and prove a useful formula for the covariance between two
interaction contrasts.

\begin{lemma}\label{lemm:variance.decomposition}
  Consider the interaction constrats in
  \eqref{eq:interaction-contrast} for a function $f$ of
  $W_1,\dots,W_K$ that are independent. Let $W'$ be an independent and
  identically distributed copy of $W$. For any disjoint $\calS, \calS'
  \subseteq [K]$, we have
  \begin{align*}
    \cov\left(I_{\calS}(W, W'), I_{\calS'}(W, W')\right)
    &= \left(-\frac{1}{2}\right)^{|\calS| +
      |\calS'|}\var\left(I_{\calS \cup \calS'}(W, W')\right) = (-1)^{|\calS| + |\calS'|} \overline{\sigma}^2_{\calS \cup \calS'}.
  \end{align*}
\end{lemma}
\begin{proof}
  For any $\calT \subseteq [K]$, we use $T_{\calT}$ to denote the mapping
  from one $K$-variate function to another that swaps $W_k$ with $W_k'$
  for all $k \in \calT$; for example,
  \[T_{\{1,2\}} (Y(W_1', W_2') - Y(W_1,
    W_2)) = Y(W_1, W_2) - Y(W_1', W_2').
  \]
  It is obvious that $T_{\calT} \circ T_{\calT} = I$, where $I$ denotes
  the identity mapping. Then
  \begin{align}
    I_{\calS \cup \calS'}(W, W')
    &=  (-1)^{|\calS'|}\sum_{\calT' \subseteq \calS'} (-1)^{|\calT'|}
      T_{\calT'}(I_{\calS}(W, W')) \notag \\
    &=  (-1)^{|\calS'|}\sum_{\calT' \subseteq \calS'} (-1)^{|\calT'|}
      I_{\calS}((W'_{\calT'}, W_{-\calT'}), (W_{\calT'},
      W'_{-\calT'})),\quad \calS \cap \calS' = \emptyset, \label{proof:eq:I}
    \\ \label{proof:eq:II}
    T_{\calT}(I_{\calS}(W, W'))
    &= (-1)^{|\calT|} I_{\calS}(W, W'), \quad \calT \subseteq \calS \subseteq [K].
  \end{align}
  Since $W'$ is an independent copy of $W$, $W$ consists of independent
  factors,
  \begingroup \allowdisplaybreaks
  \begin{align*}
    &\quad~\cov\left(I_{\calS}(W, W'), I_{\calS'}(W, W')\right)\\
    &=\left(\frac{1}{2}\right)^{|\calS'|}\cov\left(I_{\calS}(W, W'),
      \sum_{\calT' \subseteq \calS'} (-1)^{|\calT'|}
      T_{\calT'}(I_{\calS'}(W, W'))\right) \tag{by Eq.~\eqref{proof:eq:II}}\\
    &= \left(\frac{1}{2}\right)^{|\calS'|}  \sum_{\calT' \subseteq \calS'} \cov\left( (-1)^{|\calT'|} T_{\{\calT'\}}(I_{S}(W, W')), I_{S'}(W, W') \right) \tag{$\Cov$ is bilinear,~$T_{\{\calT'\}} \circ T_{\{\calT'\}} = I$}\\
    &= \left(\frac{1}{2}\right)^{|\calS'|}   \cov\left(\sum_{\calT' \subseteq \calS'} (-1)^{|\calT'|} T_{\{\calT'\}}(I_{S}(W, W')), I_{S'}(W, W') \right) \tag{$\Cov$ is bilinear}\\
    &= \left(-\frac{1}{2}\right)^{|\calS'|}  \cov\left(I_{S \cup S'}(W, W')), I_{S'}(W, W') \right)  \tag{Eq.~\eqref{proof:eq:I}}\\
    &=  \left(-\frac{1}{2}\right)^{|\calS'|}  \cdot \left(\frac{1}{2}\right)^{|\calS|}  \sum_{\calT \subseteq \calS} \cov\left( (-1)^{|\calT|} T_{\calT}(I_{\calS \cup \calS'}(W, W')), I_{S'}(W, W') \right) \tag{Eq.~\eqref{proof:eq:II}}\\
    &=  \left(-\frac{1}{2}\right)^{|\calS'|}  \cdot \left(\frac{1}{2}\right)^{|\calS|}  \sum_{\calT \subseteq \calS} \cov\left( I_{\calS \cup \calS'}(W, W'), (-1)^{|\calT|} T_{\calT}(I_{S'}(W, W')) \right) \tag{$\Cov$ is bilinear,~$T_{\{\calT\}} \circ T_{\{\calT\}} = I$}\\
    &=  \left(-\frac{1}{2}\right)^{|\calS'|}  \cdot \left(\frac{1}{2}\right)^{|\calS|}  \cov\left(I_{S \cup \calS'}(W, W'), \sum_{\calT \subseteq \calS} (-1)^{|\calT|} T_{\calT}(I_{S'}(W, W')) \right) \tag{$\Cov$ is bilinear}\\
    &= \left(- \frac{1}{2} \right)^{|\calS'|+|\calS|}
      \cov\left(I_{\calS \cup \calS'}(W, W'), I_{\calS \cup
      \calS'}(W, W')\right) \tag{Eq.~\eqref{proof:eq:I}} \\
    &= \left(-\frac{1}{2}\right)^{|\calS| +
      |\calS'|}\var\left(I_{\calS \cup \calS'}(W, W')\right).
  \end{align*}
  \endgroup
  The second equality in the Lemma follows from  \eqref{eq:superset-pick-freeze}.
\end{proof}

Because $\calE(W)$ is a finite algebra, there exists a unique signed measure $\xi_3$ that is consistent with the specification of the total explainability $\xi_3(\vee_{k \in \calS} W_k)$ for all $\calS \subseteq [K]$ as in \eqref{eq:xi-sobol-upper}. In particular, we can first define $\xi_3(\wedge_{k \in \calS} W_k)$ for all $\calS \subseteq [K]$ using the inclusion-exclusion principle:
\begin{equation} \label{eq:xi3-interaction}
  \xi_3(\wedge_{k \in \calS} W_k) = \sum_{\calS' \subseteq \calS}
  (-1)^{|\calS'|+1} \cdot \xi_3(\vee_{k \in \calS'} W_k).
\end{equation}
we can then construct a signed measure $\xi_3$ on $\calE(W)$ by defining the
explainabilities of the atoms of $\calE(W)$: for any $\calS \subset [K]$, recursively define
\begin{equation} \label{eq:xi3-atom}
  \xi_3 \left((\wedge_{k \in \calS} W_k) \wedge (\wedge_{k \notin
      \calS} \neg W_k) \right) = \xi_3(\wedge_{k \in \calS} W_k) -
  \sum_{\calS' \supset \calS} \xi_3 \left((\wedge_{k \in \calS'} W_k)
    \wedge (\wedge_{k \notin \calS'} \neg W_k) \right).
\end{equation}
It is easy to see that
\[
  \xi_3(\vee{k \in [K]} W_k) = \frac{\Var(f(W) - f(W'))}{2 \Var(f(W))} = 1.
\]
So to prove $\xi_3$ is a probability measure, it suffices to show that the explainability of any atom is non-negative, i.e.
\begin{equation} \label{eq:non-negative-xi}
  \xi_3 \left((\wedge_{k \in \calS} W_k) \wedge (\wedge_{k \notin
      \calS} \neg W_k) \right) \geq 0, \quad \text{for all}~\calS
  \subseteq [K]
\end{equation}
Before we prove \eqref{eq:non-negative-xi}, we first show that the interaction explainability defined by \eqref{eq:xi3-interaction} is consistent with \eqref{eq:xi-superset}.

\begin{lemma}
  For any $\calS \subseteq [K]$ we have
  \begin{equation}
    \label{eq:xi-3-interaction}
    \xi_3(\wedge_{k \in \calS} W_k) = \frac{\var[I_{\calS} (W,
      W')]}{2^{|\calS|}\var(f(W))} = \xi_4(\wedge_{k \in \calS} W_k).
  \end{equation}
\end{lemma}
\begin{proof}
  The second equality in \eqref{eq:xi-3-interaction} follows from
  \eqref{eq:superset-pick-freeze} and \eqref{eq:xi-superset}. To prove
  the first equality, we first prove the following identity by
  induction on $|\calS|$:
  \begin{equation}
    \label{eq:inclusion-exclusion-pick-freeze}
    \var\left(Y(W) - Y(W'_{\calS}, W_{-\calS})\right) =
    \sum_{\emptyset \neq \calS' \subseteq \calS}
    \left(-\frac{1}{2}\right)^{|\calS'|-1} \var\left(I_{\calS'}(W,
      W')\right).
  \end{equation}
  The base case $|\calS| = 1$ (so $\calS = \{k\}$) is the
  definition of $\xi_3(W_k)$. Next, suppose
  \eqref{eq:inclusion-exclusion-pick-freeze} holds for any $\calS$ of
  cardinality $K-1$. Now, consider $\calS$ with $K$ factors, and let
  $\calS' \subset \calS$ where $|\calS'| = |\calS| - 1$, we have
  \begin{align*}
    &\quad~\var\left(Y(W) - Y(W'_{\calS}, W_{-\calS})\right) \\
    &= \var\left(\left(Y(W) - Y(W_{\calS'}', W_{-\calS'})\right)\right) + \var\left(Y(W_{\calS'}', W_{-\calS'}) - Y(W_{\calS}', W_{-\calS})\right) \\
    &\quad~+ 2 \cov\left(Y(W) - Y(W_{\calS'}', W_{-\calS'}), Y(W_{\calS'}', W_{-\calS'}) - Y(W_{\calS}', W_{-\calS})\right)\\
    &= \sum_{\emptyset \neq \calS'' \subseteq \calS'}
      \left(-\frac{1}{2}\right)^{|\calS''|-1} \var\left(I_{ \calS''}(W, W')\right) \tag*{(Induction hypothesis)}\\
    &\quad + \var\left(I_{\calS \setminus \calS'}(W, W')\right) 
    + 2 \cov\Big(Y(W'_{\calS'}, W_{-\calS'}) - Y(W), -I_{\calS \setminus
    \calS'}(W, W') \Big) \tag*{(Swap~$W_{\calS'}, W_{\calS'}'$)}
    \\
    &= \sum_{\emptyset \neq \calS'' \subseteq \calS'}
      \left(-\frac{1}{2}\right)^{|\calS''|-1} \var\left(I_{ \calS''}(W, W')\right) + \var\left(I_{\calS \setminus \calS'}(W, W')\right) \\
    &\quad~ - 2 \cov\Big(\sum_{\emptyset \neq \calS'' \subseteq \calS'} I_{\calS''}(W, W'), ~I_{\calS \setminus \calS'}(W, W') \Big). \quad \tag*{(Anchored decomposition)}
    \\
    &= \sum_{\emptyset \neq \calS'' \subseteq \calS'}
      \left(-\frac{1}{2}\right)^{|\calS''|-1} \var\left(I_{ \calS''}(W, W')\right)
      + \var\left(I_{\calS \setminus \calS'}(W, W')\right) \\
    &\quad~- 2 \sum_{\emptyset \neq \calS'' \subseteq \calS'}
      \left(-\frac{1}{2}\right)^{|\calS''|+1}\var\Big(I_{\calS'' \cup
      (\calS \setminus \calS')}(W, W') \Big) \tag*{(Lemma \ref{lemm:variance.decomposition})}\\
    &=  \sum_{\emptyset \neq \calS'' \subseteq \calS}
      \left(-\frac{1}{2}\right)^{|\calS''|-1} \var\left(I_{\calS''}(W, W')\right).
  \end{align*}
  This establishes
  \eqref{eq:inclusion-exclusion-pick-freeze}. The first equality in
  \eqref{eq:xi-3-interaction} then follows from applying the
  inclusion-exclusion principle.
\end{proof}

We next prove \eqref{eq:non-negative-xi}
by induction on the number of explanatory variables $K$. The $K = 1$
case is trivial. Now we assume \eqref{eq:non-negative-xi} is true up
to $K-1$. Note
that $\wedge_{k \notin \calS} \neg W_k = \neg (\vee_{k \notin \calS} W_k)$,
so if $|\calS| < K-1$ we can view $(\vee_{k \notin \calS} W_k)$ as one ``large'' factor
and apply the inductive hypothesis (the atom $(\wedge_{k \in \calS} W_k) \wedge (\wedge_{k \notin
  \calS} \neg W_k)$ in $\calE(W_1,\dots,W_K)$ can be viewed as an atom in a smaller explanation algebra, and the definition of total explainability in these two algebras are identical).
It is left to show \eqref{eq:non-negative-xi} for when $|\calS| = K-1$. Without loss of generality, suppose $\calS = [K-1]$. By \eqref{eq:xi3-atom},
\begin{align*}
  \xi((\wedge_{k \in \calS} W_k) \wedge \neg W_K)
  &= \xi(\wedge_{k \in \calS} W_k) - \xi(\wedge_{k \in \calS \cup \{K\}} W_k) 
  \\
  &=  \frac{1}{2^{|\calS|}} \var(I_{\calS}(W, W')) - \frac{1}{2^{|\calS|+1}} \var(I_{\calS \cup \{K\}}(W, W')). \tag*{}
\end{align*}
By Eq.~\eqref{proof:eq:I},
\begin{align*}
  &\quad~\var(I_{\calS \cup \{K\}}(W, W'))\\
  &= \var(I_{\calS}(W, W')) + \var(T_{\{K\}}(I_{\calS}(W, W'))) - 2 \cov(T_{\{K\}}(I_{\calS}(W, W')), I_{\calS}(W, W'))) \\
  &= 2 \var(I_{\calS}(W, W')) - 2 \E[T_{\{K\}}(I_{\calS}(W, W')) I_{\calS}(W, W')],
\end{align*}
where we use $\E[I_{\calS}(W, W')] = 0$ in the last equation. So
\begin{align*}
  &\xi((\wedge_{k \in \calS} W_k) \wedge \neg W_K) \\
  =& \frac{1}{2^{|\calS|}} \E[T_{\{K\}}(I_{\calS}(W, W')) I_{\calS}(W, W')] \\
  =& \frac{1}{2^{|\calS|}} \E\left[\E[T_{\{K\}}(I_{\calS}(W, W')) I_{\calS}(W, W') \mid W_{-K}, W'_{-K}] \right]  \\
  =& \frac{1}{2^{|\calS|}} \E\left[\E[T_{\{K\}}(I_{\calS}(W, W')) \mid W_{-K}, W'_{-K}] \cdot \E[I_{\calS}(W, W') \mid W_{-K}, W'_{-K}] \right]\\
  =& \frac{1}{2^{|\calS|}} \E\left[\E^2[I_{\calS}(W, W') \mid W_{-K}, W'_{-K}]\right] \\
  \ge& 0,
\end{align*}
where the third equality follows from the conditional independence $W_K \independent W_K' \mid W_{-K}, W'_{-K}$.

\section{Additional results for the {US Income}
  dataset}\label{appe:sec:figure}

\subsection{Exploratory data analysis}\label{sec:distribution}

\begin{figure}[tbp]
  \centering
  \includegraphics[width = 0.7\textwidth]{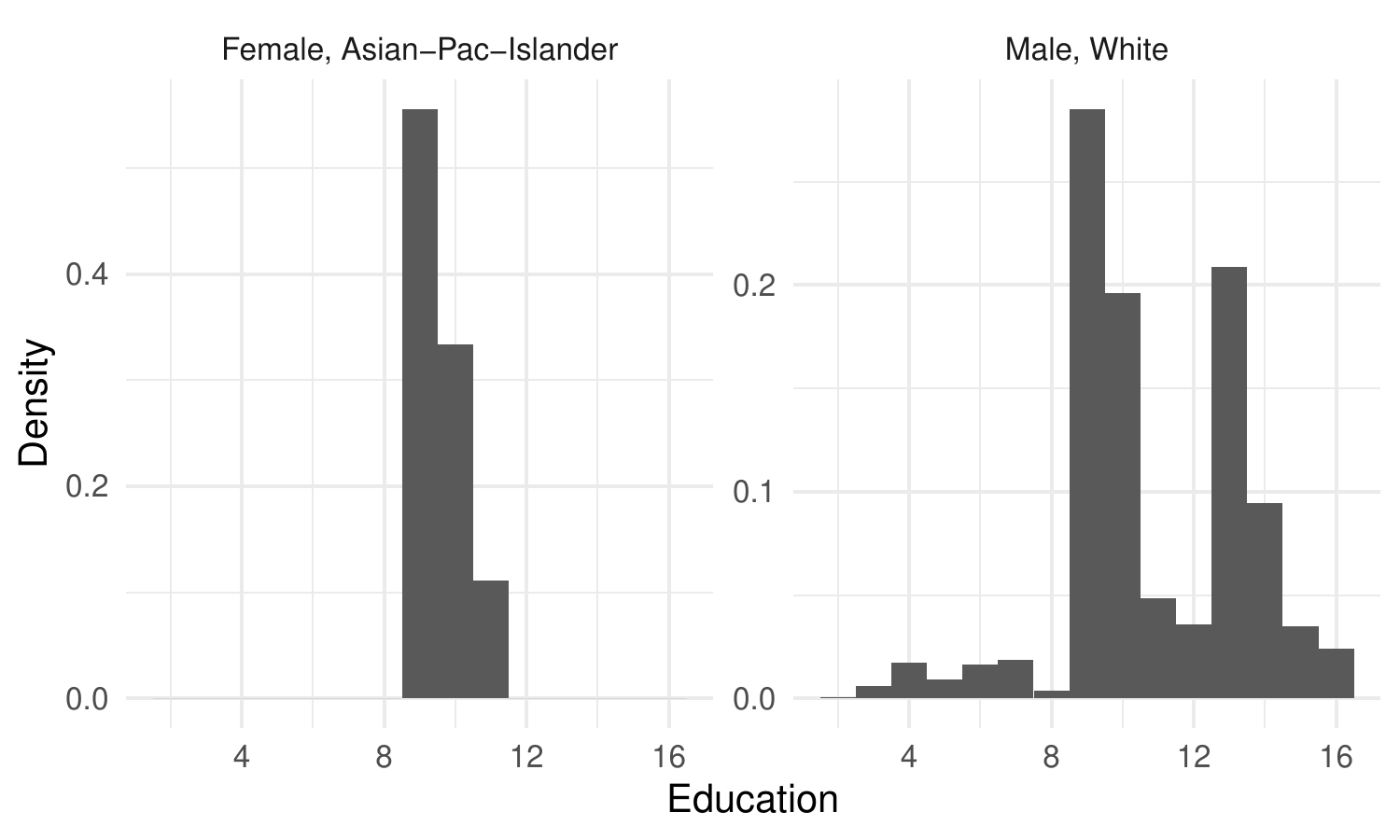}
  \caption{Education distributions conditional on sex, race in age group [45, 50). Histograms for Female, Asian–Pac–Islander (left) and Male, White (right). Shapes are non-Gaussian and differ across groups.}
  \label{fig:education.residual}
\end{figure}

\Cref{fig:education.residual} shows education distributions (years of schooling) in age group [45, 50) for two sex, race configurations: Female, Asian–Pac–Islander and Male, White.
Both histograms are clearly non-Gaussian, with visible spikes at schooling milestones (9/10 for both configurations, and also 13 for Male, White).
The shapes differ across groups, highlighting heterogeneity that motivates using estimation methods without assuming normal or homogeneous errors.

\begin{figure}[tbp]
  \centering
  \includegraphics[width = 0.7\textwidth]{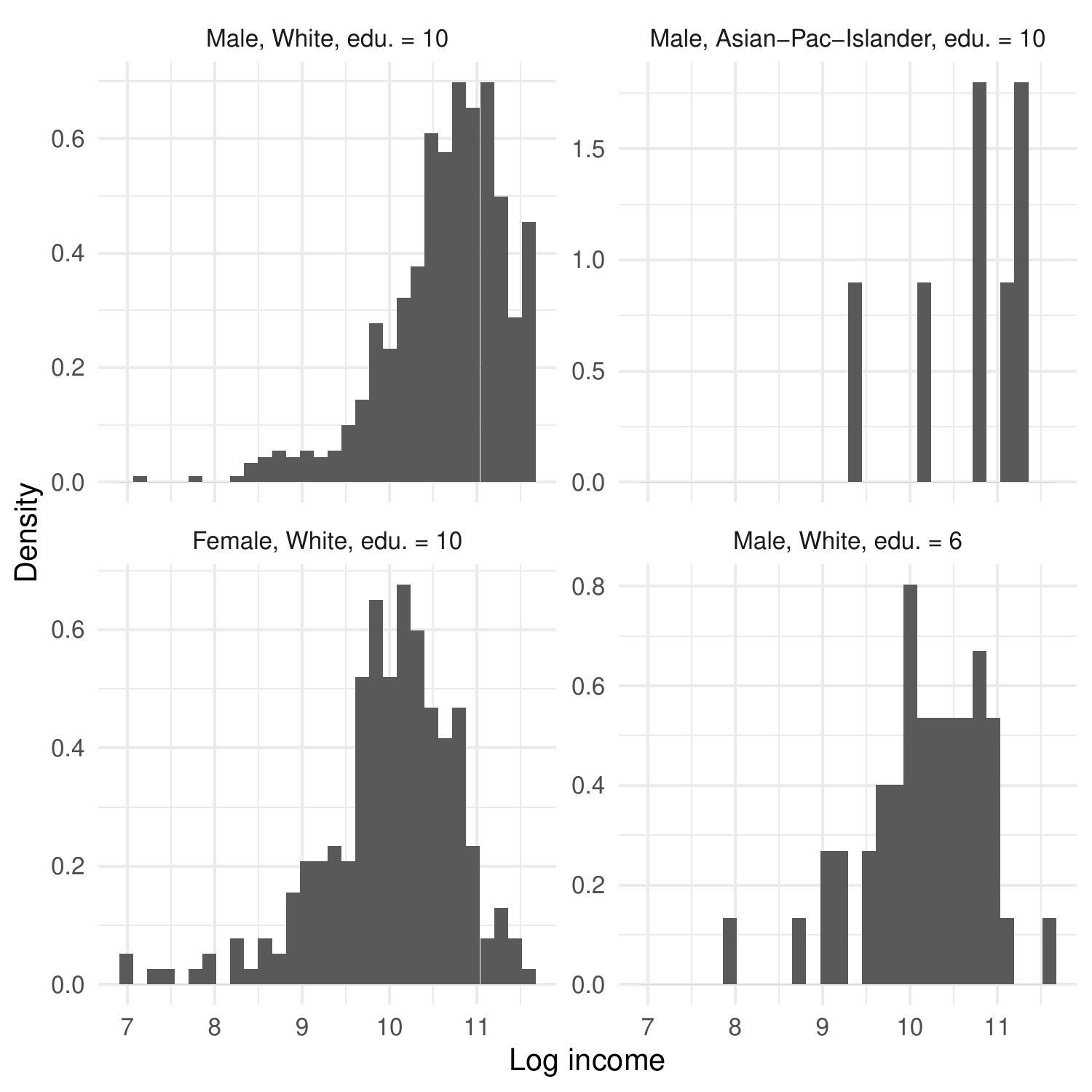}
  \caption{Log of income distributions conditional on sex, race, education in age group [45, 50). Histograms for Male, White, education of value $10$ (top left), Male, Asian–Pac–Islander, education of value $10$ (top right), Female, White, education of value $10$ (bottom left), Male, White, education of value $6$ (bottom right). Shapes are non-Gaussian and differ across groups.}
  \label{fig:income.residual}
\end{figure}

Similarly, \Cref{fig:income.residual} plots log-income distributions in age group [45, 50) for four specific sex, race, education configurations.
All four cases show skewed, non-Gaussian shapes.
The differences in both shape and tail indicate distribution heterogeneity.

\subsection{Alternative estimation method}\label{sec:other.estimation}

We describe the two alternative estimation procedures. Once the conditional
distributions for all nodes are specified, the {sampling} and
{estimation} steps (including the ``pick-freeze'' coupling) are exactly as
in \Cref{sec:estimation}; these methods differ only in how they estimate the
conditional distributions of education and income.

\paragraph{Additive independent noise model.}
For education and income we posit an additive–noise model with independent errors.
We use two–fold cross–fitting to
avoid overfitting: split the sample into two folds; on fold 1, fit the
conditional mean with XGBoost and compute residuals on fold 2; then swap the two folds and repeat (fit on fold 2, compute residuals on fold 1).
We then pool the out–of–fold residuals from two folds and use their empirical distribution as the estimated error distribution.

\paragraph{Heteroskedastic Gaussian noise model.}
For this estimator, education and income are assumed conditionally Gaussian with parent-dependent variance.
We again use two-fold cross-fitting: split the sample, fit the conditional mean with XGBoost on one fold and compute out-of-fold residuals on the other, then swap folds.
To learn the variance function, we regress a transformed residual:  the log of the absolute residual squared on the parent features using a second XGBoost model. Before taking logs, we apply a small floor to the absolute residual to avoid numerical issues and instability. Finally, we exponentiate back to obtain the estimated variance.

\begin{figure}[t]
  \centering
  \begin{minipage}{1\textwidth}
    \centering
    \includegraphics[clip, trim = 0cm 0cm 0cm 0cm, width = 1\textwidth]{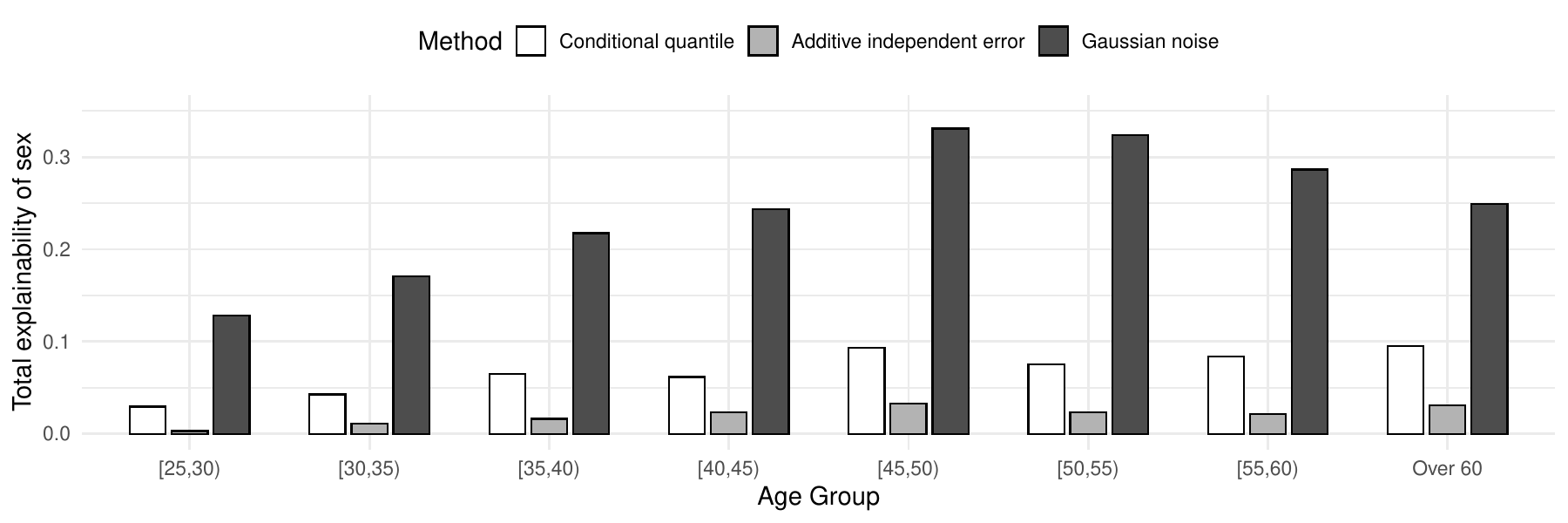}
    \caption*{(a) Total explainability of sex}
  \end{minipage}\\
  \begin{minipage}{1\textwidth}
    \centering
    \includegraphics[clip, trim = 0cm 0cm 0cm 0cm, width = 1\textwidth]{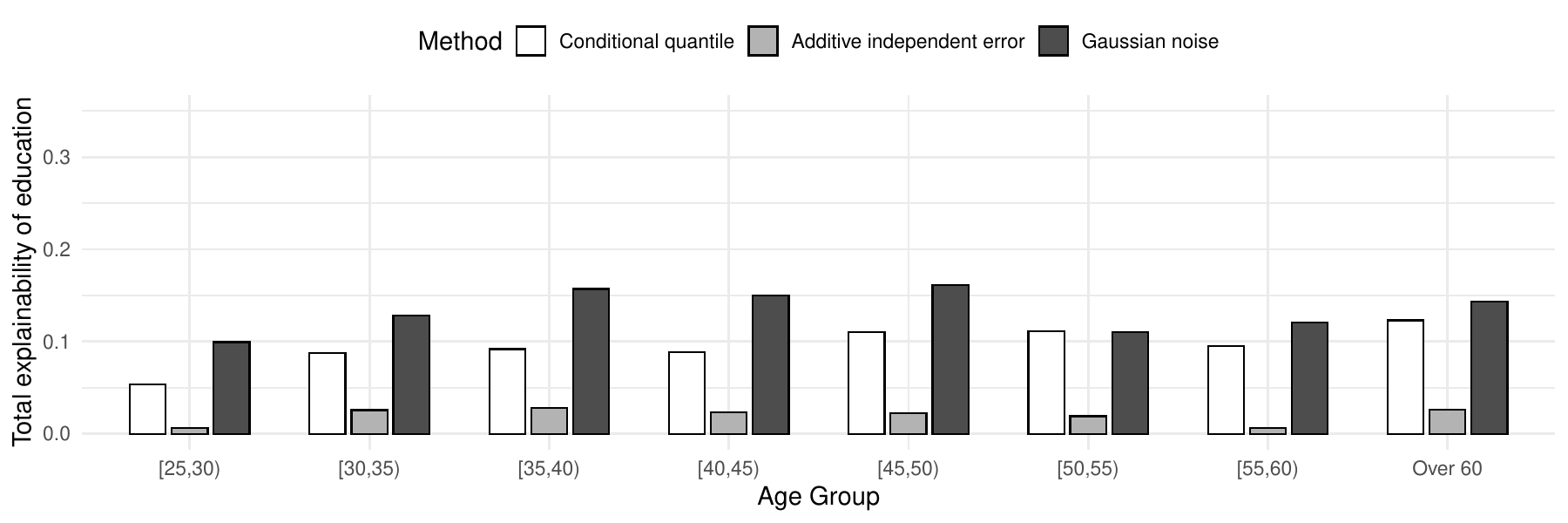}
    \caption*{(b) Total explainability of education}
  \end{minipage}
  \caption{Comparison of the estimated total explainability for
    sex or education across age groups, using different estimation methods.} \label{fig:other.methods}
\end{figure}

In \Cref{fig:other.methods}, we compare the estimated total explainability for
sex or education across age groups, using different estimation methods.
The additive independent–noise model systematically {underestimates} total explainability because it fails to
account for the dependence of the error distribution on the inputs. For example, consider education, the estimated total explainabilities by the additive independent noise model-based procedure can be as low as $20\%$ of those obtained from the conditional–quantile procedure.
In contrast, the Gaussian–noise variant substantially {underestimates the total variance of the outcome} because it fails to account for the tail behavior, thereby inflating the estimated fraction of variance explained.
For example, the total explainabilities of sex estimated by this method can reach up to three times those produced by the conditional–quantile approach.

Overall, these
discrepancies across methods highlight that the choice of estimation method has a non-negligible impact on the magnitude of the estimated explainabilities.
Our main analysis relies on the conditional–quantile estimator, which captures heteroskedasticity and non-normal error structures and is thus the most suitable method for the {US Income} data.

\subsection{Comparison with Shapley's value}
\label{sec:comp-with-shapl}

\begin{figure}[t]
  \includegraphics[clip, trim = 0cm 0cm 0cm 0cm, width = 1\textwidth]{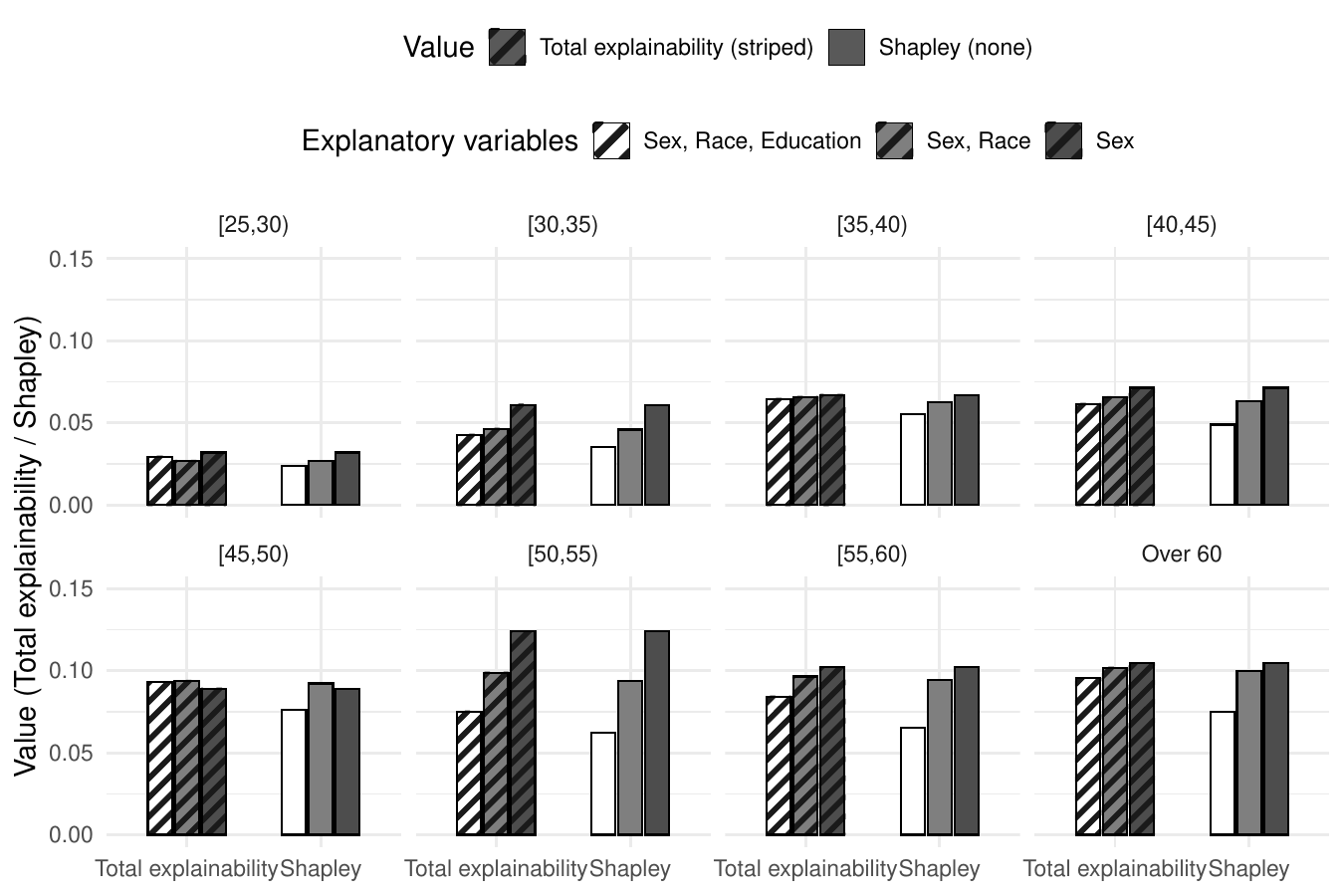}
  \caption{Comparison of the total explainability and Shapley value of
    sex.}
  \label{fig:compare-shapley}
\end{figure}

We compute the Shapley value for sex using the three different sets of
explanatory variables in \Cref{fig:real.data}, using total
explanability (i.e.\ Sobol's upper index) as the set value function. By
definition, Shapley's value depends on which other explanatory variables are
included. In this example, when sex is the only node, the Shapley value for
  sex coincides with its value function, which is set to the total
  explainability of sex. When sex, race, and education are all
  included, and the only nonzero interaction is between sex and
  education, the Shapley value for sex equals our the total
  explainability of sex minus one half of the sex-education
  interaction explainability and thus is smaller than the Shapley
  value in the sex-only case. Indeed, Shapley's value demonstrate
  considerably more instability than total counterfactual
  explainability in \Cref{fig:compare-shapley}.

\subsection{Partial identification}\label{sec:partial.ID}

When the comonotonicity assumption does not hold, point identification is no
longer possible, and we instead compute lower and upper bounds.
The values based on comonotonicity are valid lower bounds, as previously discussed.
For upper bounds, we
follow the approach of \cite{leiHeritabilityCounterfactualPerspective2025}. Noting that the upper bounds are generally not tight.

\begin{table}[t]
  \centering
  \caption{Lower and upper bounds of total explainabilities by age group.}\label{tab:partial}
  \begin{tabular}{c|cc|cc|cc}
    \toprule
    \multirow{2}{*}{Age Group} & \multicolumn{2}{c|}{Race} & \multicolumn{2}{c|}{Sex} & \multicolumn{2}{c}{Education} \\
                               & Lower & Upper & Lower & Upper & Lower & Upper \\
    \midrule
    $[25,30)$ & 0.0092 & 0.9669 & 0.0295 & 0.9734 & 0.0539 & 0.9439 \\
    $[30,35)$ & 0.0169 & 0.9396 & 0.0427 & 0.9540 & 0.0876 & 0.9038 \\
    $[35,40)$ & 0.0110 & 0.9190 & 0.0646 & 0.9600 & 0.0919 & 0.8959 \\
    $[40,45)$ & 0.0099 & 0.9196 & 0.0615 & 0.9672 & 0.0885 & 0.8999 \\
    $[45,50)$ & 0.0030 & 0.8581 & 0.0930 & 0.9772 & 0.1105 & 0.8478 \\
    $[50,55)$ & 0.0061 & 0.8895 & 0.0750 & 0.9663 & 0.1112 & 0.8748 \\
    $[55,60)$ & 0.0047 & 0.8899 & 0.0841 & 0.9812 & 0.0953 & 0.8802 \\
    $\geq 60$ & 0.0036 & 0.8903 & 0.0955 & 0.9809 & 0.1230 & 0.8811 \\
    \bottomrule
  \end{tabular}
\end{table}

In \Cref{tab:partial}, we report the lower and upper bounds of total
explainabilities for race, sex, and education across age groups. The
bounds are generally quite wide, reflecting the fundamental
uncertainty that arises when working with counterfactual quantities.

\end{document}